\documentclass[lettersize,journal]{IEEEtran}

\usepackage{amsmath,amsfonts}
\usepackage{algorithmic}
\usepackage{algorithm}
\usepackage{array}
\usepackage[caption=false,font=normalsize,labelfont=sf,textfont=sf]{subfig}
\usepackage{textcomp}
\usepackage{stfloats}
\usepackage{url}
\usepackage{verbatim}
\usepackage{graphicx}
\usepackage{cite}
\usepackage{multirow}
\usepackage{amssymb}
\usepackage{tikz}
\usepackage{bm} 
\usepackage{booktabs}

\DeclareMathOperator*{\argmin}{arg\,min}
\DeclareMathOperator*{\argmax}{arg\,max}

\begin{document}

\title{Differentiable NMS via Sinkhorn Matching for End-to-End Fabric Defect Detection}


\author{
	\IEEEauthorblockN{Zhengyang Lu\textsuperscript{1}, Bingjie Lu\textsuperscript{2}, Weifan Wang\textsuperscript{1} and Feng Wang\textsuperscript{1}}\\
	\IEEEauthorblockA{\textsuperscript{1} Jiangnan University, China\\
		\textsuperscript{2} Central South University, China\\
		luzhengyang@jiangnan.edu.cn}
}



\maketitle

\begin{abstract}
Fabric defect detection confronts two fundamental challenges. First, conventional non-maximum suppression disrupts gradient flow, which hinders genuine end-to-end learning. Second, acquiring pixel-level annotations at industrial scale is prohibitively costly. Addressing these limitations, we propose a differentiable NMS framework for fabric defect detection that achieves superior localization precision through end-to-end optimization. We reformulate NMS as a differentiable bipartite matching problem solved through the Sinkhorn-Knopp algorithm, maintaining uninterrupted gradient flow throughout the network. This approach specifically targets the irregular morphologies and ambiguous boundaries of fabric defects by integrating proposal quality, feature similarity, and spatial relationships. Our entropy-constrained mask refinement mechanism further enhances localization precision through principled uncertainty modeling. Extensive experiments on the Tianchi fabric defect dataset demonstrate significant performance improvements over existing methods while maintaining real-time speeds suitable for industrial deployment. The framework exhibits remarkable adaptability across different architectures and generalizes effectively to general object detection tasks.
\end{abstract}


\section{Introduction}

\begin{figure}[h]
	\centering
	\includegraphics[width=\linewidth]{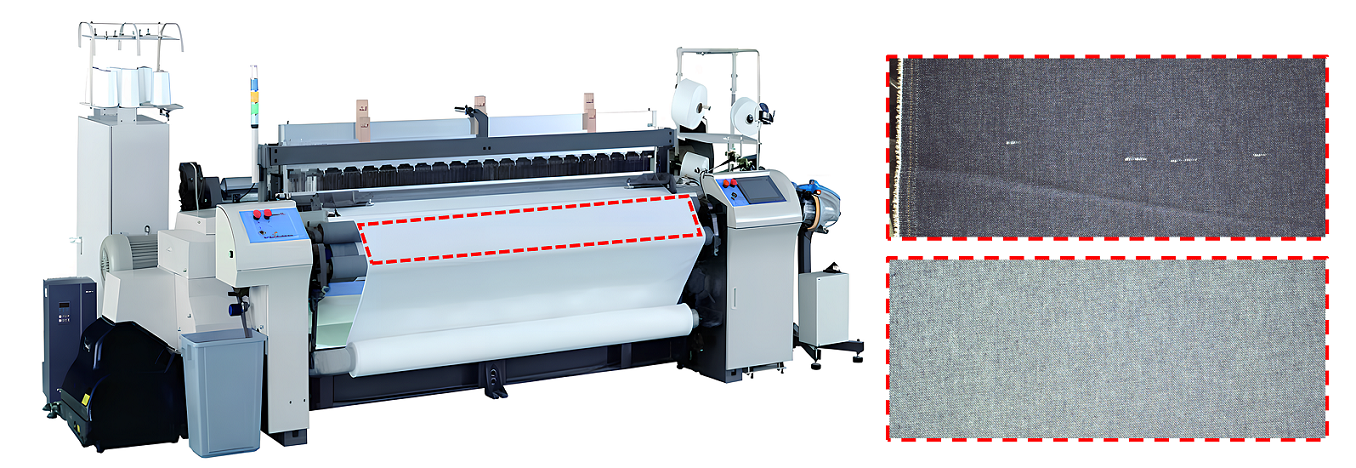}
	\caption{Textile manufacturing system with automated inspection zones (highlighted in red) and sample fabric outputs showing pattern variations relevant for defect detection.}
	\label{fig:machine}
\end{figure}

Fabric defect detection constitutes a critical quality control step in textile manufacturing, with even minor flaws potentially leading to significant economic losses and customer dissatisfaction. Traditional inspection methods rely on human visual examination, which suffers from inconsistency, fatigue, and inefficiency when faced with the high-speed, continuous production environments characteristic of modern textile mills. Figure \ref{fig:machine} illustrates a typical textile manufacturing system with integrated inspection zones used for automated defect detection. While deep learning-based object detection approaches have shown promise for automating this task, they face unique challenges in the textile domain: subtle defect boundaries, complex background textures, diverse defect morphologies, and the prohibitive cost of obtaining pixel-level annotations in industrial settings.

Despite recent advances in convolutional neural networks for industrial defect detection, existing approaches predominantly employ traditional non-maximum suppression (NMS) as a post-processing step to eliminate redundant bounding boxes. This greedy selection mechanism, while computationally efficient, fundamentally limits detection performance through its non-differentiable nature, breaking gradient flow during back-propagation and preventing true end-to-end optimization of the detection pipeline. In fabric defect scenarios, where defects often present irregular shapes and ambiguous boundaries against complex textile backgrounds, the hard thresholding in standard NMS can prematurely eliminate valid proposals or retain suboptimal ones, significantly impacting detection precision and recall. These limitations become particularly pronounced when dealing with elongated defects like broken yarns or scratches that span significant image regions.

The proposed method offers a pragmatic alternative to traditional detection methods by leveraging more readily available box-level labels rather than expensive pixel-level annotations\cite{xia2022cbash}. However, existing weakly supervised detection methods struggle with the inherent noise in proposal masks generated from classification attention maps, often producing imprecise localization results that fail to capture the subtle fabric defect boundaries.

This paper introduces a novel end-to-end fabric defect detection framework that addresses these fundamental limitations through differentiable NMS via Hungarian matching. Our approach reformulates the traditional non-maximum suppression operation as a bipartite matching problem between observed proposals and latent defect regions, solved through the differentiable Sinkhorn-Knopp algorithm. This formulation maintains uninterrupted gradient flow throughout the network while providing theoretical convergence guarantees, enabling joint optimization of proposal generation, selection, and refinement stages without relying on heuristic post-processing steps. Unlike previous attempts at differentiable NMS that focus solely on operations differentiability, our method integrates proposal quality, feature similarity, and spatial relationships within a unified matching framework that inherently addresses the ambiguous boundaries and irregular morphologies characteristic of fabric defects. The transformation from cost matrix to soft assignment matrix enables precise control over proposal suppression while preserving gradient information critical for end-to-end training. This holistic approach allows the model to leverage complementary cues from different components, significantly improving detection performance on complex textile patterns.

To address the inherent noise in proposal masks, we further introduce an entropy-constrained mask refinement mechanism that optimizes proposal probabilities while maintaining sufficient uncertainty through principled entropy constraints. This is accomplished via a Frank-Wolfe optimization procedure with provable convergence properties, which progressively refines the quality of defect masks by balancing confidence with uncertainty in a theoretically grounded manner. Our entropy-constrained approach prevents overconfident but incorrect predictions during early training, while gradually converging toward more precise localization as training progresses. Additionally, we incorporate spatial coherence regularization to enforce locally consistent predictions, which is particularly beneficial for detecting elongated fabric defects like broken yarns and scratches that span significant image regions with varying intensity.

Extensive experiments on the challenging Tianchi fabric defect dataset demonstrate that our DNMS-FDD framework significantly outperforms existing methods, achieving state-of-the-art performance across multiple evaluation metrics while maintaining computational efficiency suitable for real-time industrial deployment. The proposed method not only improves detection accuracy but also shows remarkable adaptability across different backbone architectures and generalization to the general object detection domain, as evidenced by competitive results on the COCO benchmark.

Within the our work, the contributions can be summarized as follows:
\begin{itemize}
	\item We reformulate NMS as a differentiable bipartite matching problem using the Sinkhorn-Knopp algorithm, enabling end-to-end training with uninterrupted gradient flow throughout the detection pipeline.
	\item We introduce an entropy-constrained mask refinement mechanism that progressively improves proposal masks through principled uncertainty modeling with theoretical convergence guarantees.
	\item We develop a unified framework that achieves competitive performance with fully supervised methods, demonstrating superior efficiency-accuracy trade-offs for industrial deployment.
\end{itemize}

\section{Related Works}

\subsection{Defect Detection Methods}

Surface defect detection has evolved substantially from conventional image processing techniques to contemporary deep learning paradigms. Early works relied on handcrafted features extracted via Gabor filters \cite{arivazhagan2006fault}, wavelet transforms \cite{ngan2011automated}, and gray-level co-occurrence matrices \cite{haralick1973textural}. These classical approaches, while computationally efficient, struggled with illumination changes and background texture variations that characterize real-world industrial environments.

Convolutional neural networks revolutionized industrial inspection by automatically learning hierarchical features from raw images. Pioneering efforts by Masci et al. \cite{masci2012steel} demonstrated CNN superiority over traditional methods in steel defect classification tasks. Subsequently, region-based architectures including Faster R-CNN \cite{ren2016faster} and RetinaNet \cite{lin2017focal} were adapted for industrial defect detection, achieving remarkable performance improvements in semiconductor \cite{lee2017convolutional}, solar panel \cite{wieler2007weakly}, traffic object\cite{shi2023fixated}, and metallic surface \cite{xie2020ffcnn} inspection. 
Self-supervised learning has emerged as a promising direction for industrial defect detection, addressing the persistent challenge of limited labeled data. 
Bergmann et al. \cite{bergmann2020uninformed} introduced a patch-based approach that leverages structural similarity to enhance autoencoder performance for unsupervised defect segmentation. Zhang et al. \cite{zhang2022adaptive} introduced memory-augmented normality learning that explicitly models the manifold of normal appearance variations. The integration of foundation models has further accelerated progress, with Gu et al. \cite{gu2024anomalygpt} demonstrating that vision-language models pre-trained on diverse internet-scale data can effectively transfer to defect detection tasks through minimal prompt engineering. Concurrently, diffusion models have shown remarkable capability in anomaly detection by learning to reconstruct normal patterns and identifying deviations during inference \cite{wyatt2022anoddpm}.

Fabric defect detection presents distinctive challenges due to complex backgrounds, fabric texture variations, and irregular defect morphologies compared to more structured industrial surfaces. Jia et al.\cite{jia2020fabric} address automated patterned fabric inspection by segmenting repetitive texture into lattice primitives and detecting anomalies through statistical comparison against templates learned from defect-free samples using a flexible multi-feature framework. With the advent of deep networks, Shao et al. \cite{shao2022pixel} leverage multitask mean teacher frameworks to achieve better performance with minimal labeled samples, addressing the annotation scarcity in industrial textile inspection applications. Recent unsupervised approaches utilize frequency-based features \cite{wu2022self} and boundary-guided anomaly synthesis \cite{chen2024progressive}, eliminating annotation requirements while achieving state-of-the-art performance on industrial datasets. Building upon fusion-based approaches, Liu et al. \cite{liu2022fabric} proposed low-rank decomposition techniques incorporating structural constraints to distinguish physical defects from normal patterns. To enhance foreground-background feature discrimination, feature contrast interference suppression \cite{wang2025enhanced} addresses background sensitivity by employing contrastive learning and knowledge distillation. Recent advances in data selection methods like PRISM \cite{bi2025prism, bi2024visual} have also shown promise for efficient training of multimodal defect detection systems.

\subsection{Non-maximum Suppression}

Non-maximum suppression (NMS) constitutes a critical post-processing stage in object detection pipelines that eliminates redundant bounding boxes through greedy selection based on confidence scores and overlap calculations. Despite the effectiveness, traditional NMS suffers from several limitations: 1) the non-differentiable nature breaks gradient flow during backpropagation, preventing end-to-end optimization; 2) hard thresholding mechanism potentially discards valuable information; and 3) the sequential processing creates computational bottlenecks in parallel architectures.

Recent advances have focused on developing differentiable alternatives to traditional NMS. Hosang et al. \cite{hosang2017learning} pioneered this direction by framing NMS as a network component trained with a combination of detection and duplicate removal losses. Building upon this, soft-NMS \cite{bodla2017soft} replaced hard thresholding with score decay functions proportional to overlap. More recent approaches have reformulated NMS as various optimization problems. He et al. \cite{he2019bounding} introduced adaptive NMS that dynamically adjusts IoU thresholds based on object density. Salscheider \cite{salscheider2021featurenms} proposed FeatureNMS which leverages semantic similarities alongside spatial overlap. Furthermore, Sun et al. \cite{sun2021sparse} presented Sparse R-CNN with learnable proposal boxes that eliminate the need for NMS entirely.

Latest researches in NMS have evolved beyond traditional greedy approaches to address the inherent limitations. Parallel implementations have significantly improved computational efficiency on embedded GPU architectures, demonstrating more than 40 times speedups while maintaining detection accuracy \cite{oro2022work}. Neural attention-driven techniques have emerged to better handle occlusions in pedestrian detection by jointly processing geometric and visual properties through sequence-to-sequence formulations \cite{symeonidis2023neural}. For overlapping object scenarios, adaptive NMS strategies dynamically adjust IoU thresholds based on object density, substantially enhancing performance for weed detection applications \cite{al2023adaptive}.

\section{Methodology}

\begin{figure*}[htb]
	\centering
	\includegraphics[width=.98\linewidth]{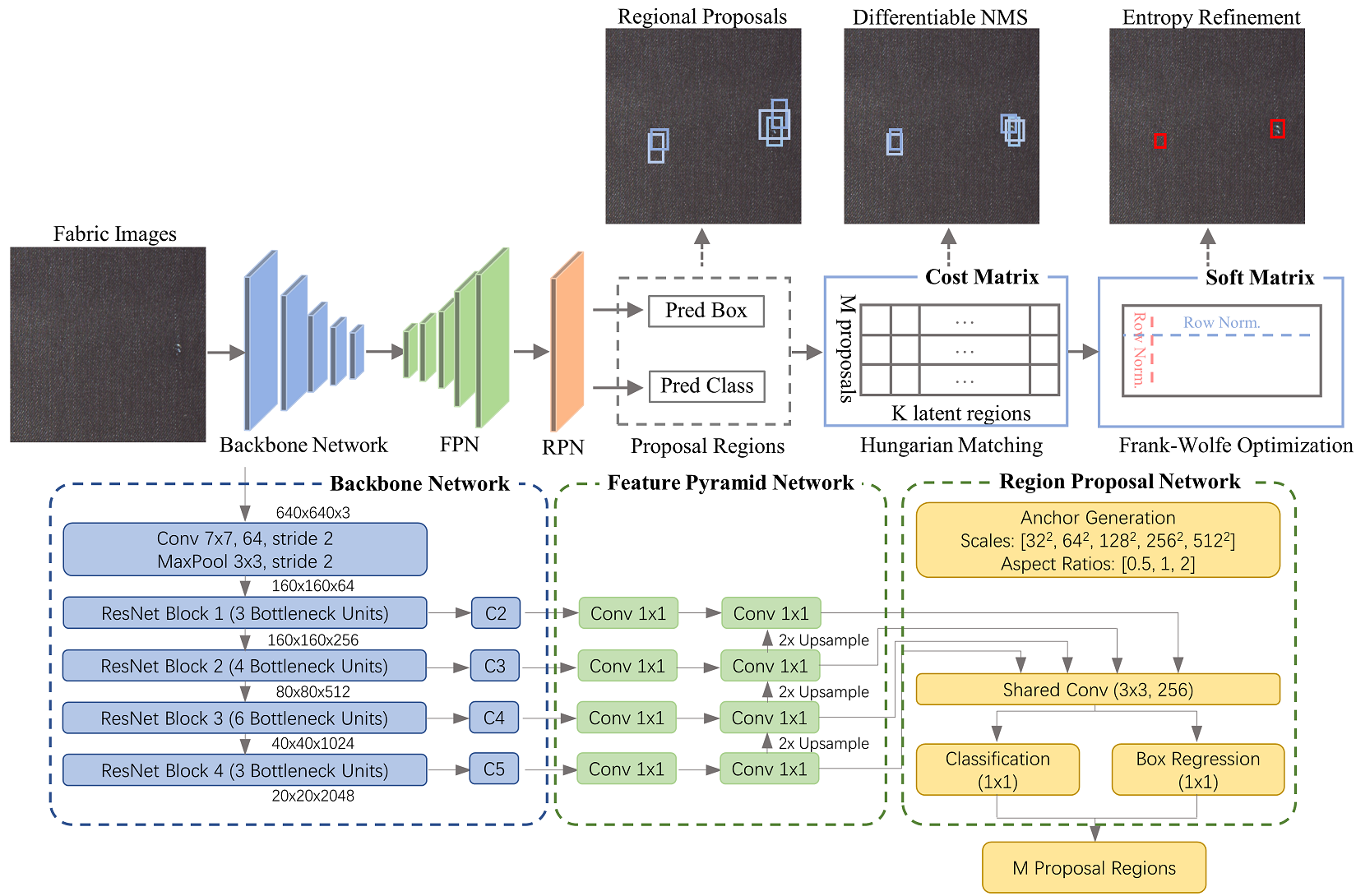}
	\caption{End-to-end fabric defect detection architecture. Top: Complete pipeline showing fabric image input through proposal generation, differentiable NMS via Hungarian matching, and entropy-constrained refinement. Bottom: Detailed network components including ResNet backbone, Feature Pyramid Network with lateral connections, and Region Proposal Network with scale-specific anchors.}
	\label{fig:full_pipeline}
\end{figure*}

Classic object detection approaches with non-differentiable NMS break gradient flow during backpropagation. Our method addresses this by reformulating NMS as a bipartite matching problem solved through the Sinkhorn-Knopp algorithm, ensuring full differentiability with theoretical convergence guarantees. Figure \ref{fig:full_pipeline} illustrates our end-to-end fabric defect detection framework, integrating a ResNet-50 backbone with Feature Pyramid Network and Region Proposal Network.

The architecture integrates a ResNet-50 backbone with Feature Pyramid Network (FPN) and Region Proposal Network (RPN) in a fully differentiable pipline. The backbone extracts hierarchical features at four scales, with resolution decreasing and channel dimension increasing progressively from C2 to C5. The FPN performs two critical functions: (1) lateral connections through 1$\times$1 convolutions normalize channel dimensions to 256 across all levels, and (2) top-down pathways with 2$\times$ upsampling operations enrich lower-level features with high-level semantics. This structure creates a scale-invariant feature representation particularly beneficial for detecting fabric defects of varying sizes. The RPN then processes each scale independently with scale-specific anchors while utilizing shared convolutional weights, maintaining computational efficiency. Unlike traditional object detection pipelines that employ non-differentiable NMS as post-processing, our architecture preserves gradient flow through all stages, enabling end-to-end optimization of the entire detection process.

\subsection{Problem Formulation}
Let $\mathcal{D} = \{(I_i, B_i)\}_{i=1}^N$ denote our training dataset, where $I_i \in \mathbb{R}^{H \times W \times 3}$ represents an input image and $B_i$ represents the corresponding bounding box annotations for defects present in the image. This approach efficiently utilizes these annotations while maintaining high detection precision.  

We formulate defect detection as a proposal generation and refinement problem. For each image $I_i$, we first generate a set of $M$ defect proposals $\mathcal{P}_i = \{p_{i,j}\}_{j=1}^M$, where each proposal $p_{i,j}$ consists of a defect confidence score $s_{i,j} \in [0, 1]$ and a spatial mask $m_{i,j} \in \{0, 1\}^{H \times W}$. Traditional methods apply non-differentiable Non-Maximum Suppression (NMS) to filter redundant proposals, which breaks gradient flow during backpropagation. Our approach introduces a differentiable alternative that maintains end-to-end trainability while addressing the inherent noise in proposal masks.

\subsection{Differentiable NMS via Hungarian Matching}
\label{subsec:dnms}

\begin{figure}[t]
	\centering
	\includegraphics[width=\linewidth]{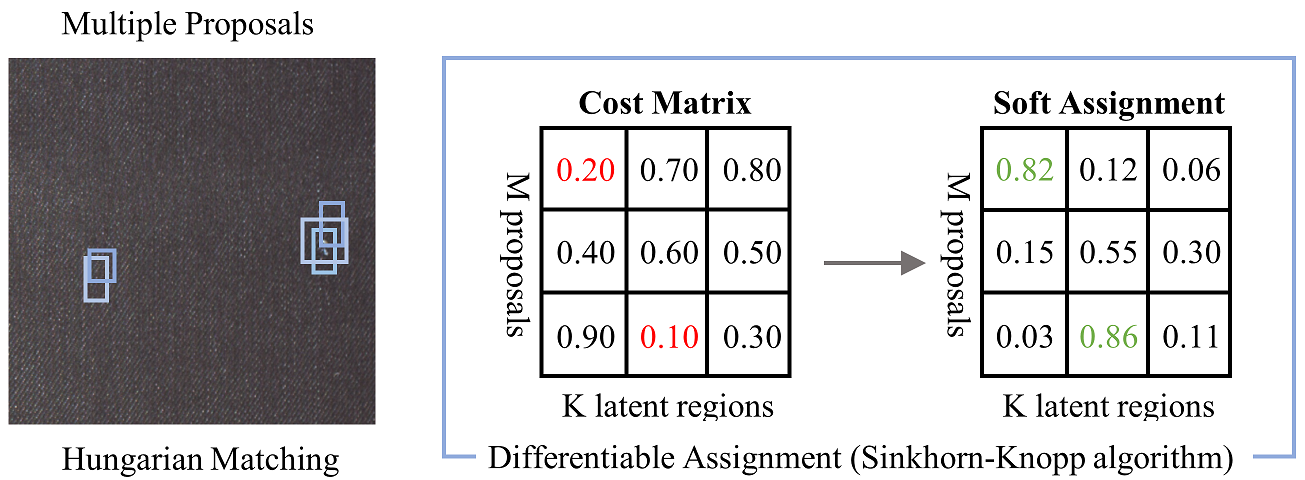}
	\caption{Detailed illustration of the differentiable NMS via Hungarian matching. Multiple proposal regions on a fabric image (left) are matched to latent defect regions through a differentiable assignment process (right). The cost matrix represents assignment costs between proposals and latent regions, with lower values (in red) indicating preferred assignments. The Sinkhorn-Knopp algorithm transforms this into a differentiable soft assignment matrix where higher values (in green) indicate stronger assignments.}
	\label{fig:hungarian_matching}
\end{figure}

Classical NMS selects proposals based on greedy confidence thresholding and overlap removal, a process that cannot be directly differentiated. We reformulate this selection process as a bipartite matching problem between observed proposals and latent defect regions. Figure \ref{fig:hungarian_matching} illustrates this process, showing how our differentiable assignment converts the cost matrix to a soft assignment matrix through the Sinkhorn-Knopp algorithm, enabling end-to-end gradient flow while effectively suppressing redundant proposals.

Given a set of $M$ proposals $\mathcal{P} = \{p_j\}_{j=1}^M$ and $K$ latent defect regions (where $K$ is estimated adaptively based on image complexity), we construct a cost matrix $\mathbf{C} \in \mathbb{R}^{M \times K}$ where each element $C_{j,k}$ represents the assignment cost between proposal $p_j$ and latent defect region $k$:

\begin{equation}
	C_{j,k} = \alpha \cdot (1 - s_j) + \beta \cdot d_{\text{feat}}(p_j, \mu_k) + \gamma \cdot d_{\text{spatial}}(p_j, \nu_k)
\end{equation}
where $s_j$ is the confidence score of proposal $p_j$, $\mu_k$ and $\nu_k$ represent the feature centroid and spatial centroid of latent defect region $k$, respectively. The functions $d_{\text{feat}}$ and $d_{\text{spatial}}$ measure feature dissimilarity and spatial distance, while $\alpha$, $\beta$, and $\gamma$ are balancing hyperparameters.

The standard Hungarian algorithm finds the optimal assignment matrix $\mathbf{P} \in \{0, 1\}^{M \times K}$ that minimizes the total assignment cost:

\begin{equation}
	\mathbf{P}^* = \argmin_{\mathbf{P}} \sum_{j=1}^M \sum_{k=1}^K C_{j,k} P_{j,k}
\end{equation}
subject to $\sum_{j=1}^M P_{j,k} \leq 1$ and $\sum_{k=1}^K P_{j,k} \leq 1$, ensuring that each proposal is assigned to at most one defect region and vice versa.

However, this discrete optimization is non-differentiable. We adopt a continuous relaxation using the Sinkhorn-Knopp algorithm to make it differentiable. We introduce a doubly-stochastic matrix $\mathbf{S} \in [0, 1]^{M \times K}$ where each element $S_{j,k}$ represents the soft assignment probability between proposal $p_j$ and latent defect region $k$.

The Sinkhorn-Knopp algorithm iteratively normalizes rows and columns of the matrix $\mathbf{A} = \exp(-\mathbf{C}/\tau)$, where $\tau$ is a temperature parameter controlling the softness of the assignment:

\begin{align}
	\mathbf{R}^{(t)} &= \mathbf{A} / (\mathbf{A}\mathbf{1}_K\mathbf{1}_K^T) \\
	\mathbf{S}^{(t)} &= \mathbf{R}^{(t)} / (\mathbf{1}_M\mathbf{1}_M^T\mathbf{R}^{(t)})
\end{align}

As $t \rightarrow \infty$ and $\tau \rightarrow 0$, $\mathbf{S}^{(t)}$ approaches a discrete permutation matrix. In practice, we run a fixed number of iterations $T$ and maintain a moderate temperature to ensure gradient flow.

The differentiable matching loss is defined as:

\begin{equation}
	\mathcal{L}_{\text{match}} = \sum_{j=1}^M \sum_{k=1}^K C_{j,k} S_{j,k} + \lambda \cdot D_{\text{KL}}(\mathbf{S} \| \mathbf{P}^*)
\end{equation}
where $D_{\text{KL}}$ is the Kullback-Leibler divergence guiding the soft assignment $\mathbf{S}$ toward the optimal discrete assignment $\mathbf{P}^*$, and $\lambda$ is a regularization parameter.

The complete procedure is summarized in Algorithm \ref{alg:dnms}. Our differentiable NMS yields a set of refined proposals $\mathcal{P}' = \{p'_j\}_{j=1}^{K}$ corresponding to the $K$ latent defect regions, by aggregating the proposals assigned to each region:

\begin{equation}
	p'_k = \sum_{j=1}^M S_{j,k} \cdot p_j
\end{equation}

This formulation effectively suppresses redundant proposals while maintaining differentiability, enabling end-to-end training of the entire detection pipeline.

\begin{algorithm}
	\caption{Differentiable NMS via Hungarian Matching}
	\label{alg:dnms}
	\begin{algorithmic}[1]
		\REQUIRE Proposals $\mathcal{P} = \{(s_j, m_j)\}_{j=1}^M$, Number of latent defect regions $K$
		\ENSURE Refined proposals $\mathcal{P}' = \{p'_k\}_{k=1}^K$
		
		\STATE Initialize centroid features $\{\mu_k\}_{k=1}^K$ and spatial centers $\{\nu_k\}_{k=1}^K$ (via k-means)
		\STATE Compute cost matrix $C_{j,k} = \alpha \cdot (1 - s_j) + \beta \cdot d_{\text{feat}}(p_j, \mu_k) + \gamma \cdot d_{\text{spatial}}(p_j, \nu_k)$
		\STATE $\mathbf{A} \leftarrow \exp(-\mathbf{C}/\tau)$ \COMMENT{Apply temperature scaling}
		
		\STATE $\mathbf{S} \leftarrow \mathbf{A}$ \COMMENT{Initialize soft assignment matrix}
		\FOR{$t = 1$ to $T$}
		\STATE $\mathbf{S} \leftarrow \mathbf{S} / (\mathbf{S}\mathbf{1}_K\mathbf{1}_K^T)$ \COMMENT{Row normalization}
		\STATE $\mathbf{S} \leftarrow \mathbf{S} / (\mathbf{1}_M\mathbf{1}_M^T\mathbf{S})$ \COMMENT{Column normalization}
		\ENDFOR
		
		\FOR{$k = 1$ to $K$}
		\STATE $p'_k \leftarrow \sum_{j=1}^M S_{j,k} \cdot p_j$ \COMMENT{Aggregate assigned proposals}
		\ENDFOR
		
		\RETURN $\mathcal{P}' = \{p'_k\}_{k=1}^K$
	\end{algorithmic}
\end{algorithm}

\subsection{Entropy-Constrained Mask Refinement}
\label{subsec:entropy}

The proposals refined by differentiable NMS can benefit from further refinement to address potential noise. We introduce an entropy-constrained optimization framework to further refine these masks while avoiding overconfident but incorrect predictions.

For a given image with proposals $\mathcal{P}'=\{p'_k\}_{k=1}^K$, we define a probability vector $\mathbf{p} \in [0, 1]^K$ where each element $p_k$ represents the probability that the $k$-th proposal contains a true defect. The goal is to find the optimal probability distribution that maximizes agreement with ground truth annotations while maintaining sufficient uncertainty through an entropy constraint:

\begin{align}
	\max_{\mathbf{p}} \quad & \sum_{k=1}^K p_k \cdot q_k\\
	\text{s.t.} \quad & \sum_{k=1}^K p_k = 1\\
	& H(\mathbf{p}) \geq \tau\\
	& 0 \leq p_k \leq 1, \quad \forall k \in \{1,2,\ldots,K\}
\end{align}
where $q_k$ is the quality score of proposal $p'_k$, computed as the maximum IoU between the proposal and any ground truth bounding box in $B_i$. This directly incorporates box-level supervision into our optimization framework. $H(\mathbf{p}) = -\sum_{k=1}^K p_k \log p_k$ is the entropy of the probability distribution, and $\tau$ is a threshold controlling the minimum required uncertainty.

This constrained optimization problem is non-trivial due to the entropy constraint. The Frank-Wolfe algorithm iteratively linearizes the objective function and solves the resulting linear program over the feasible region. For our entropy-constrained optimization, we proceed as follows:

Let $f(\mathbf{p}) = \sum_{k=1}^K p_k \cdot q_k$ be our objective function, and $\mathcal{F} = \{\mathbf{p} \in [0, 1]^K : \sum_{k=1}^K p_k = 1, H(\mathbf{p}) \geq \tau\}$ be the feasible region. At iteration $t$, we:

1. Compute the gradient $\nabla f(\mathbf{p}^{(t)}) = [q_1, q_2, \ldots, q_K]^T$.

2. Find the solution to the linearized problem:
\begin{align}
	\mathbf{s}^{(t)} = \argmax_{\mathbf{s} \in \mathcal{F}} \langle \nabla f(\mathbf{p}^{(t)}), \mathbf{s} \rangle
\end{align}

3. Update the solution:
\begin{align}
	\mathbf{p}^{(t+1)} = (1 - \gamma_t) \mathbf{p}^{(t)} + \gamma_t \mathbf{s}^{(t)}
\end{align}
where $\gamma_t \in [0, 1]$ is the step size.

The challenge lies in solving the linearized problem in step 2, as it involves an entropy constraint. We handle this by considering the dual problem:

\begin{align}
	\max_{\mathbf{s}} \min_{\lambda \geq 0} \sum_{k=1}^K s_k q_k - \lambda \cdot (H(\mathbf{s}) - \tau)
\end{align}

For a fixed $\lambda$, the optimal $\mathbf{s}$ can be derived analytically:

\begin{align}
	s_k = \frac{\exp(q_k/\lambda)}{\sum_{j=1}^K \exp(q_j/\lambda)}
\end{align}

We then perform binary search on $\lambda$ to find the value that satisfies $H(\mathbf{s}) = \tau$ (or $H(\mathbf{s}) \geq \tau$ for the boundary case).

For the step size $\gamma_t$, we use a line search to maximize the objective:

\begin{align}
	\gamma_t = \argmax_{\gamma \in [0,1]} f((1 - \gamma) \mathbf{p}^{(t)} + \gamma \mathbf{s}^{(t)})
\end{align}

The entropy-constrained mask refinement procedure is detailed in Algorithm \ref{alg:fw}. The Frank-Wolfe algorithm guarantees convergence to the global optimum for our convex objective and constraint set, with a convergence rate of $\mathcal{O}(1/t)$.

\begin{algorithm}
	\caption{Entropy-Constrained Mask Refinement via Frank-Wolfe}
	\label{alg:fw}
	\begin{algorithmic}[1]
		\REQUIRE Proposals $\mathcal{P}' = \{p'_k\}_{k=1}^K$, Quality scores $\{q_k\}_{k=1}^K$, Entropy threshold $\tau$, Max iterations $T_{\max}$
		\ENSURE Refined probability distribution $\mathbf{p}^*$
		
		\STATE Initialize $\mathbf{p}^{(0)} = [1/K, 1/K, \ldots, 1/K]^T$ \COMMENT{Uniform distribution}
		\FOR{$t = 0$ to $T_{\max} - 1$}
		\STATE $\nabla f(\mathbf{p}^{(t)}) \leftarrow [q_1, q_2, \ldots, q_K]^T$
		
		\STATE \COMMENT{Find $\lambda$ such that $H(\mathbf{s}) \geq \tau$ via binary search}
		\STATE $\lambda_{\min} \leftarrow 0.001$, $\lambda_{\max} \leftarrow 100$
		\WHILE{$\lambda_{\max} - \lambda_{\min} > \epsilon$}
		\STATE $\lambda \leftarrow (\lambda_{\min} + \lambda_{\max}) / 2$
		\STATE $s_k \leftarrow \exp(q_k / \lambda) / \sum_{j=1}^K \exp(q_j / \lambda)$ for $k = 1, 2, \ldots, K$
		\STATE $H(\mathbf{s}) \leftarrow -\sum_{k=1}^K s_k \log s_k$
		\IF{$H(\mathbf{s}) < \tau$}
		\STATE $\lambda_{\min} \leftarrow \lambda$
		\ELSE
		\STATE $\lambda_{\max} \leftarrow \lambda$
		\ENDIF
		\ENDWHILE
		
		\STATE \COMMENT{Perform line search for optimal step size}
		\STATE $\gamma_t \leftarrow$ LineSearch($\mathbf{p}^{(t)}$, $\mathbf{s}$, $f$)
		
		\STATE $\mathbf{p}^{(t+1)} \leftarrow (1 - \gamma_t) \mathbf{p}^{(t)} + \gamma_t \mathbf{s}$
		
		\IF{$\|f(\mathbf{p}^{(t+1)}) - f(\mathbf{p}^{(t)})\| < \delta$}
		\STATE \textbf{break} \COMMENT{Early stopping if converged}
		\ENDIF
		\ENDFOR
		\RETURN $\mathbf{p}^* \leftarrow \mathbf{p}^{(t+1)}$
	\end{algorithmic}
\end{algorithm}

\subsection{Integrated End-to-End Training Framework}
\label{subsec:integrated}

The complete DNMS-FDD framework integrates the differentiable NMS with entropy-constrained refinement in an end-to-end trainable pipeline. The key innovation lies in maintaining uninterrupted gradient flow throughout the detection process, enabling joint optimization of proposal generation, selection, and refinement stages without relying on heuristic post-processing or multi-stage training paradigms that plague conventional approaches.

Algorithm \ref{alg:end2end} outlines our training procedure, which enables precise localization capabilities through end-to-end optimization. The forward pass generates initial proposals and refines them through differentiable operations that preserve gradient information, creating high-quality proposal masks. Unlike existing methods that employ separate optimization objectives for different stages, our framework uses a unified loss function that balances classification accuracy ($\mathcal{L}_{\text{cls}}$), matching quality ($\mathcal{L}_{\text{match}}$), and spatial coherence ($\mathcal{L}_{\text{reg}}$). This holistic approach allows the model to leverage complementary cues from different components, addressing potential noise in proposal generation through end-to-end optimization of confidence scores, proposal matching, and mask refinement.

The spatial coherence regularization term $\mathcal{L}_{\text{reg}}$ merits special attention as it enforces locally consistent predictions by penalizing fragmented masks. We implement this regularization through Total Variation (TV) minimization:

\begin{equation}
	\mathcal{L}_{\text{reg}} = \sum_{k=1}^{K} \sum_{i,j} |m'_{k}(i+1,j) - m'_{k}(i,j)| + |m'_{k}(i,j+1) - m'_{k}(i,j)|
\end{equation}

This penalizes high-frequency transitions in proposal masks, encouraging piecewise-smooth defect regions that preserve structural continuity across fabric anomalies.

\begin{algorithm}
	\caption{DNMS-FDD: End-to-End Training}
	\label{alg:end2end}
	\begin{algorithmic}[1]
		\REQUIRE Training dataset $\mathcal{D} = \{(I_i, B_i)\}_{i=1}^N$, learning rate $\eta$, temperature $\tau$, entropy threshold $\tau_H$, balancing weights $\lambda_1, \lambda_2, \lambda_3$
		\ENSURE Trained model parameters $\theta$
		
		\STATE Initialize network parameters $\theta$ randomly or from pretrained model
		\FOR{each epoch}
		\FOR{each minibatch $\{(I_b, B_b)\}_{b=1}^B \subset \mathcal{D}$}
		\STATE Extract features $F_b = f_{\theta}(I_b)$ using backbone network
		\STATE Generate initial proposals:\\$\mathcal{P}_b = \{(s_{b,j}, m_{b,j})\}_{j=1}^M$ from $F_b$
		\STATE Estimate number of latent defect regions $K_b$
		\STATE Apply Differentiable NMS (Algorithm \ref{alg:dnms}):\\
		$\mathcal{P}'_b = \text{DNMS}(\mathcal{P}_b, K_b, \tau)$
		\STATE Apply entropy-constrained mask refinement \\(Algorithm \ref{alg:fw}):\\
		$\mathbf{p}^*_b = \text{MaskRefine}(\mathcal{P}'_b, \{q_{b,k}\}_{k=1}^{K_b}, \tau_H)$
		\STATE Compute classification loss:\\ $\mathcal{L}_{\text{cls}} = \frac{1}{B}\sum_{b=1}^B \text{BCE}(\mathbf{p}^*_b, \mathbf{y}_b)$
		\STATE Compute matching loss: \\$\mathcal{L}_{\text{match}} = \frac{1}{B}\sum_{b=1}^B \sum_{j=1}^M \sum_{k=1}^{K_b} C_{b,j,k} S_{b,j,k}$
		\STATE Compute spatial coherence regularization:\\
		$\mathcal{L}_{\text{reg}} = \frac{1}{B}\sum_{b=1}^B \sum_{k=1}^{K_b} \text{TV}(m'_{b,k})$
		\STATE Compute total loss: $\mathcal{L} = \mathcal{L}_{\text{cls}} + \lambda_1 \mathcal{L}_{\text{match}} + \lambda_2 \mathcal{L}_{\text{reg}}$
		\STATE Compute gradients: $\nabla_{\theta} \mathcal{L}$
		\STATE Update parameters: $\theta \leftarrow \theta - \eta \nabla_{\theta} \mathcal{L}$
		\ENDFOR
		\STATE Adjust learning rate according to schedule
		\ENDFOR
		\RETURN $\theta$
	\end{algorithmic}
\end{algorithm}

\subsection{Convergence Guarantees}

Our framework integrates two optimization algorithms with well-established convergence properties. For the Frank-Wolfe algorithm in entropy-constrained mask refinement, we analyze convergence through the duality gap framework. Given our linear objective $f(\mathbf{p}) = \langle \mathbf{q}, \mathbf{p} \rangle$ and convex feasible region $\mathcal{F}$ with diameter $D$, the duality gap at iteration $t$ is defined as:
\begin{align}
	g(\mathbf{p}^{(t)}) = \max_{\mathbf{s} \in \mathcal{F}} \langle \nabla f(\mathbf{p}^{(t)}), \mathbf{p}^{(t)} - \mathbf{s} \rangle
\end{align}

The optimality gap after $T$ iterations is bounded by:
\begin{align}
	f(\mathbf{p}^*) - f(\mathbf{p}^{(T)}) \leq g(\mathbf{p}^{(T)}) \leq \frac{LD^2}{T+2}
\end{align}
where $L$ is the Lipschitz constant of $\nabla f$. Since our entropy constraint forms a convex set and $f$ is linear with $L=0$, the bound simplifies to $\mathcal{O}(1/T)$ convergence.

For the Sinkhorn algorithm in differentiable NMS, the convergence rate is governed by:
\begin{align}
	\|\mathbf{P}^{(t)} - \mathbf{P}^*\|_F \leq C \cdot \rho^t
\end{align}
where $\rho = \tanh(\kappa/(4\tau))$ with $\kappa$ determined by the cost matrix structure. This establishes linear convergence at rate $\mathcal{O}(\rho^t)$, with a trade-off between convergence speed and gradient flow controlled by temperature parameter $\tau$. A comprehensive convergence analysis with complete proofs is provided in Appendix.

\section{Experimental Results}

\subsection{Tianchi Fabric dataset}

\begin{figure*}[t]
	\centering
	\subfloat[Defect Distribution]{\includegraphics[width=0.48\linewidth]{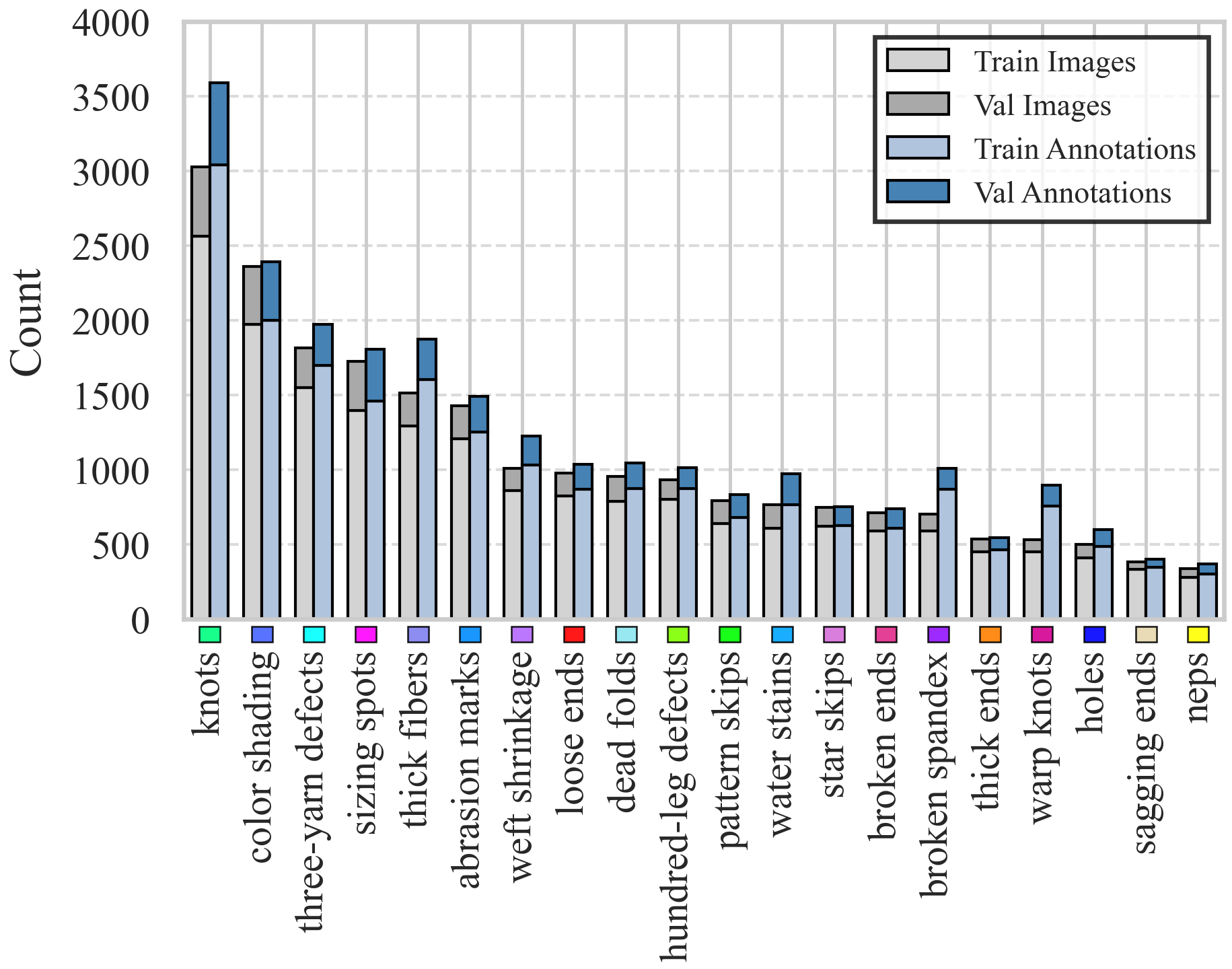}}
	\hfill
	\subfloat[Annotation Density]{\includegraphics[width=0.48\linewidth]{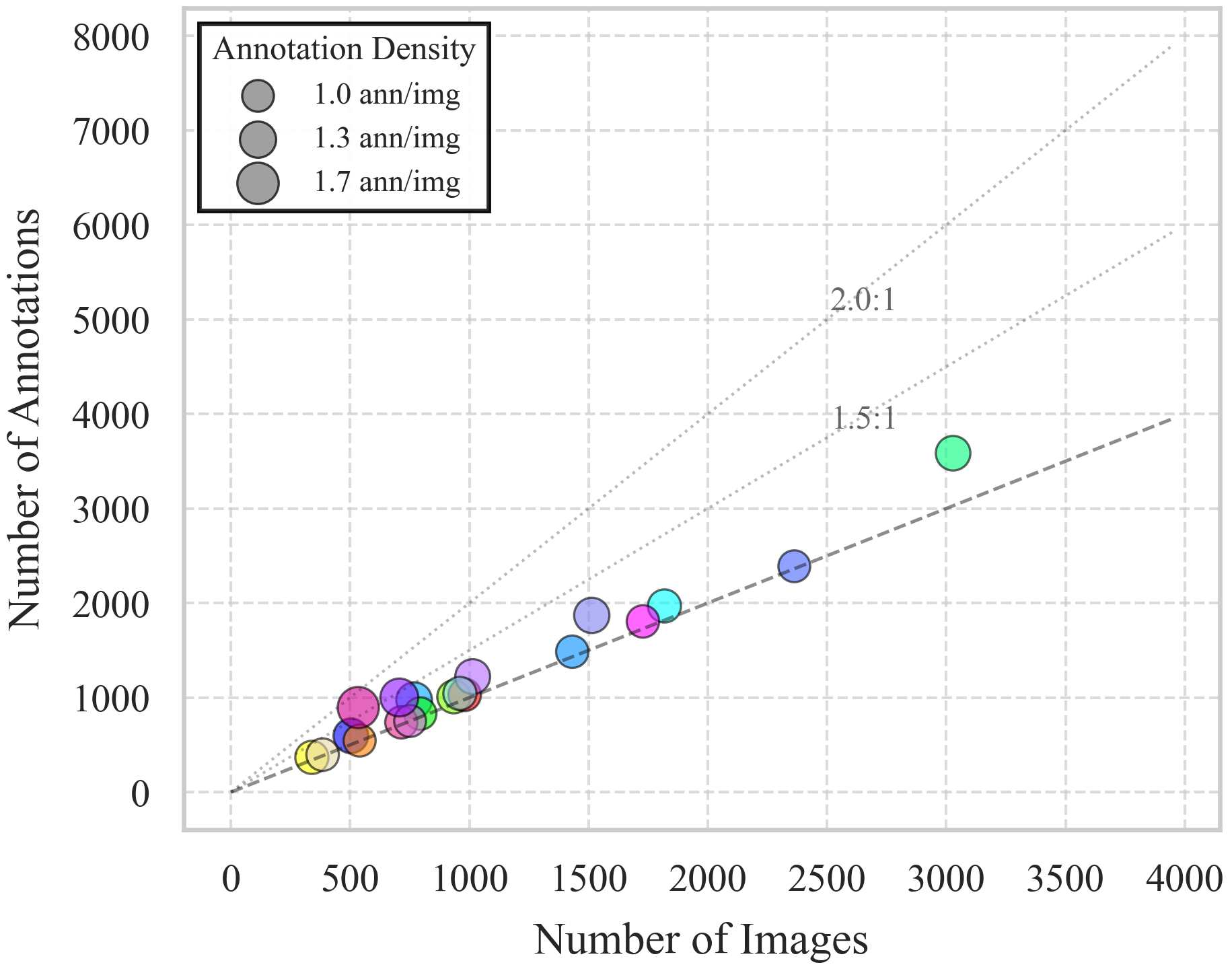}}
	\caption{Statistical analysis of the Tianchi textile defect dataset: (a) Category-wise distribution of images and annotations;  (b) Spatial annotation density visualization with bubble diameter proportional to annotations-per-image ratio.}
	\label{fig:defect_distribution}
\end{figure*}

The Tianchi Fabric dataset \cite{tianchifabric} represents a comprehensive industrial benchmark comprising 5,913 defective and 3,663 non-defective fabric images with 9,523 annotations across 20 defect categories. Original high-resolution images (2446$\times$1000 pixels) were partitioned using a 4:1 train-validation split, with each image subsequently segmented into 640$\times$640 sub-images for optimal training efficiency, yielding 17,152 training and 3,360 validation samples after removing non-defective segments.

Figure \ref{fig:defect_distribution} illustrates the dataset's distribution characteristics and annotation patterns. The category distribution in Figure \ref{fig:defect_distribution}(a) reveals substantial imbalance, with knots and color shading being predominant categories. Figure \ref{fig:defect_distribution}(b) further highlights the varying annotation densities across categories, with warp knots and broken spandex exhibiting notably higher annotations-per-image ratios. This distribution complexity presents significant challenges for detection approaches that must handle both class imbalance and varying defect densities.

\subsection{Experimental Setting}

We evaluate our DNMS-FDD framework on Tianchi fabric defect detection benchmark\cite{tianchifabric}. For fair comparison, all experiments follow standard evaluation protocols with 5-fold cross-validation, reporting mean Average Precision (mAP), F1-score, and IoU metrics. 

This work is implemented on PyTorch 1.9.0 \cite{paszke2019pytorch}, with experiments conducted on dual NVIDIA RTX A6000 GPUs with 48GB memory. We adopt ResNet-50 pretrained on ImageNet as the backbone feature extractor, followed by a Region Proposal Network that generates 256 initial proposals per image. For the differentiable Hungarian matching, we set temperature parameter $\tau=0.1$, with Sinkhorn iterations $T=10$. The entropy threshold for mask refinement is set to 0.6 with maximum Frank-Wolfe iterations $T_{max}=50$. We train the network using Adam optimizer~\cite{kingma2014adam}, weight decay 1e-6, and initial learning rate 0.01 with cosine annealing schedule for 100 epochs. Data augmentation includes random horizontal flipping. 

\subsection{Comparison with State-of-the-art Methods}

For comparison, we integrate our DNMS module with popular detection frameworks: AMFF\cite{arivazhagan2006fault}, C-RCNN~\cite{lu2023fabric}, Faster R-CNN\cite{ren2016faster}, RetinaNet\cite{lin2017focal}, FCOS\cite{tian2019fcos}, Tood~\cite{feng2021tood}, FDDA~\cite{mei2024research}, YOLOv5s~\cite{jocher2020ultralytics}, PAMF~\cite{luAnchorFreeDefectDetector2023}, YOLOX-CATD~\cite{wang2023fabric}, AFAM~\cite{wang2023adaptively}, YOLO-TTD~\cite{yue2022research}, and Deformable DETR\cite{zhu2020deformable}, demonstrating consistent improvements across architectures.

\begin{table*}[h]
	\centering
	\caption{Comprehensive Comparison with SOTA Methods on Tianchi Fabric Dataset}
	\label{table:comparison}
	\resizebox{0.75\textwidth}{!}{
		\begin{tabular}{lrrrrrrr}
			\toprule
			\multirow{2}{*}{Method} & \multicolumn{4}{c}{Detection Performance (\%)} & \multicolumn{3}{c}{Efficiency} \\
			\cmidrule(lr){2-5} \cmidrule(lr){6-8}
			& AP$_{50}$ & AP$_{75}$ & AP$_{90}$ & mAP & Params (M) & GFLOPs & FPS\\
			\midrule
			AMFF~\cite{arivazhagan2006fault} & 17.50 & 8.32 & 2.14 & 9.32 & 83.90 & 182.5 & 5.3 \\
			C-RCNN~\cite{lu2023fabric} & 19.20 & 10.45 & 3.76 & 11.14 & 81.30 & 168.4 & 5.8 \\
			Faster R-CNN~\cite{ren2016faster} & 35.90 & 23.17 & 9.84 & 22.97 & 25.60 & 94.3 & 12.5 \\
			RetinaNet~\cite{lin2017focal} & 40.25 & 27.63 & 11.45 & 26.44 & 37.74 & 88.5 & 16.2 \\
			FCOS~\cite{tian2019fcos} & 42.35 & 29.44 & 12.17 & 27.99 & 32.36 & 75.8 & 19.3 \\
			Tood~\cite{feng2021tood} & 44.10 & 31.52 & 14.05 & 29.89 & 53.26 & 65.9 & 15.7 \\
			Deformable DETR~\cite{zhu2020deformable} & 47.45 & 34.78 & 15.62 & 32.62 & 41.30 & 72.4 & 10.2 \\
			FDDA~\cite{mei2024research} & 49.80 & 36.21 & 16.44 & 34.15 & 38.52 & 68.3 & 14.8 \\
			YOLOv5s~\cite{jocher2020ultralytics} & 50.20 & 36.75 & 16.82 & 34.59 & 14.40 & 21.5 & 42.3 \\
			PAMF~\cite{luAnchorFreeDefectDetector2023} & 53.10 & 39.42 & 18.35 & 36.96 & 36.81 & 47.2 & 25.6 \\
			YOLOX-CATD~\cite{wang2023fabric} & 54.63 & 40.85 & 19.27 & 38.25 & 27.15 & 38.6 & 31.8 \\
			AFAM~\cite{wang2023adaptively} & 56.70 & 42.53 & 21.46 & 40.23 & 69.63 & 86.4 & 17.5 \\
			YOLO-TTD~\cite{yue2022research} & 56.74 & 43.18 & 22.07 & 40.67 & 24.52 & 32.1 & 36.4 \\
			\midrule
			Baseline & 57.40 & 44.25 & 23.18 & 41.61 & 11.10 & 20.4 & 48.2 \\
			DNMS-FDD (Ours) & \textbf{63.85} & \textbf{49.76} & \textbf{26.94} & \textbf{46.85} & \textbf{10.82} & \textbf{19.7} & \textbf{49.5} \\
			\bottomrule
		\end{tabular}
	}
\end{table*}

Table \ref{table:comparison} presents a comprehensive performance analysis across modern fabric defect detection methods. Our DNMS-FDD framework demonstrates substantial improvements over all comparative approaches, achieving a 5.24\% mAP gain over our baseline and 6.18\% improvement over the best competitor, YOLO-TTD. This advancement becomes particularly pronounced at higher precision thresholds, with AP$_{90}$ reaching 26.94\%, representing a 22.1\% relative improvement over YOLO-TTD. Traditional methods (AMFF, C-RCNN) struggle with the dataset's intrinsic complexities, while anchor-based approaches (Faster R-CNN, RetinaNet) establish a viable performance foundation but remain constrained by discrete NMS operations. Transformer-based Deformable DETR achieves respectable performance but falls short in computational efficiency. Notably, DNMS-FDD simultaneously improves detection accuracy while reducing computational overhead, operating at 49.5 FPS with merely 10.82M parameters and 19.7 GFLOPs, representing an optimal trade-off between precision and industrial deployment feasibility.

\begin{figure*}[h]
	\centering
	\subfloat[Structure-related fabric defects]{\includegraphics[width=0.49\textwidth]{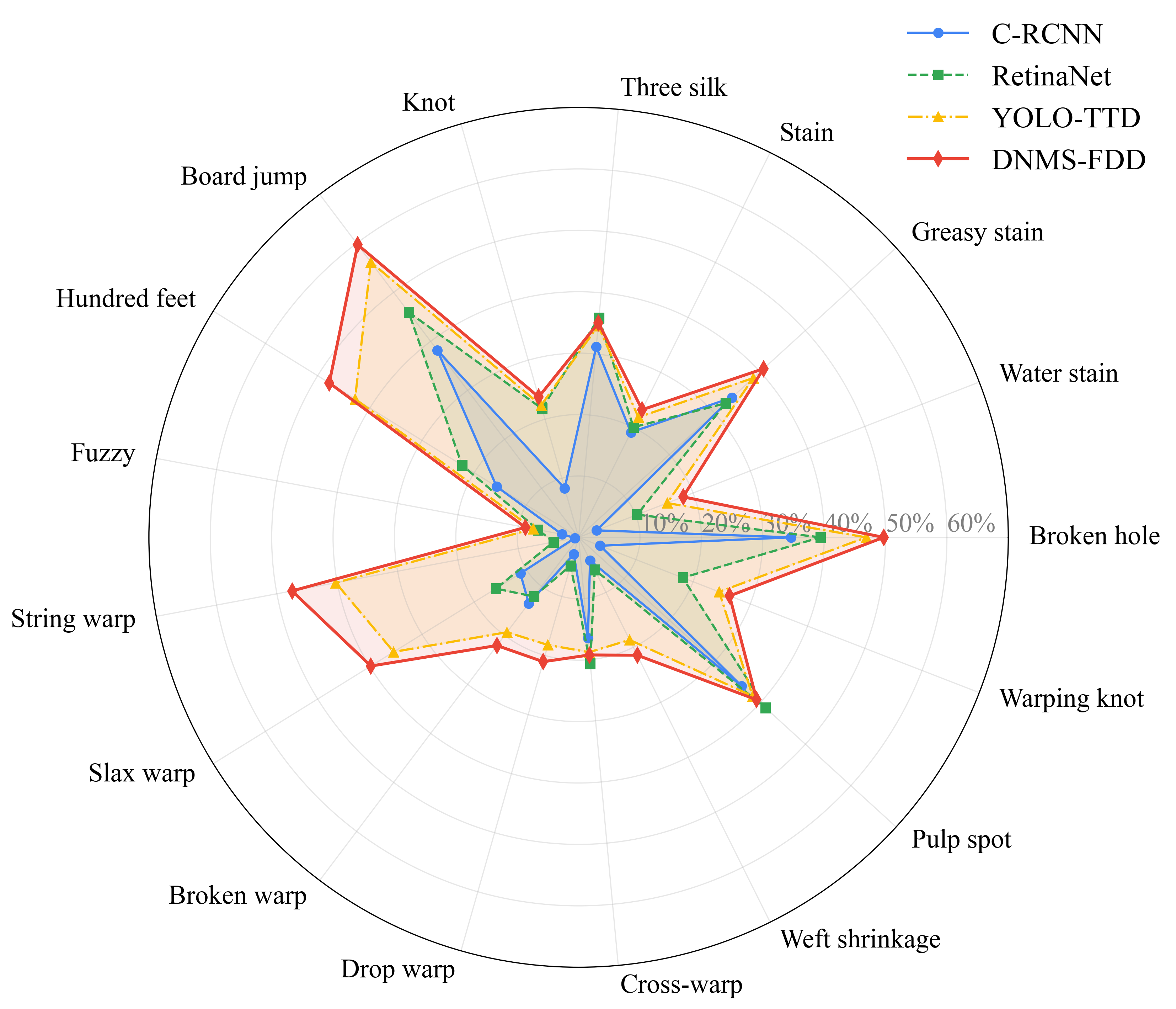}%
		\label{fig:radar_structure}}
	\hfil
	\subfloat[Surface and appearance defects]{\includegraphics[width=0.49\textwidth]{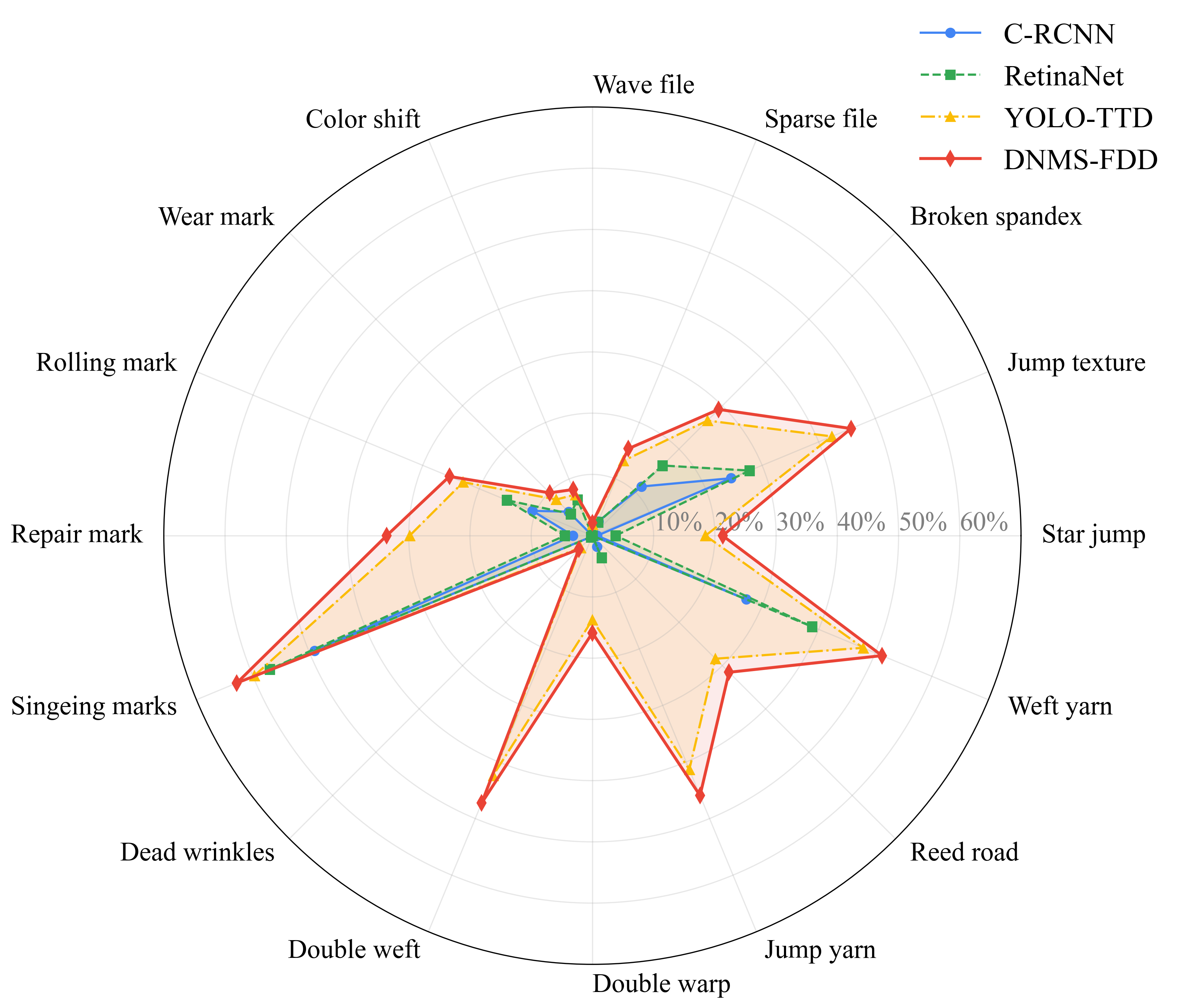}%
		\label{fig:radar_appearance}}
	\caption{Performance comparison of fabric defect detection methods across different defect categories. The radar charts illustrate mAP (\%) values for (a) structure-related defects including broken holes, water stains, and thread-related abnormalities, and (b) surface anomalies such as star jumps, repair marks, and rolling marks.}
	\label{fig:radar_comparison}
\end{figure*}

\begin{figure*}[h]
	\centering
	\includegraphics[width=0.95\linewidth]{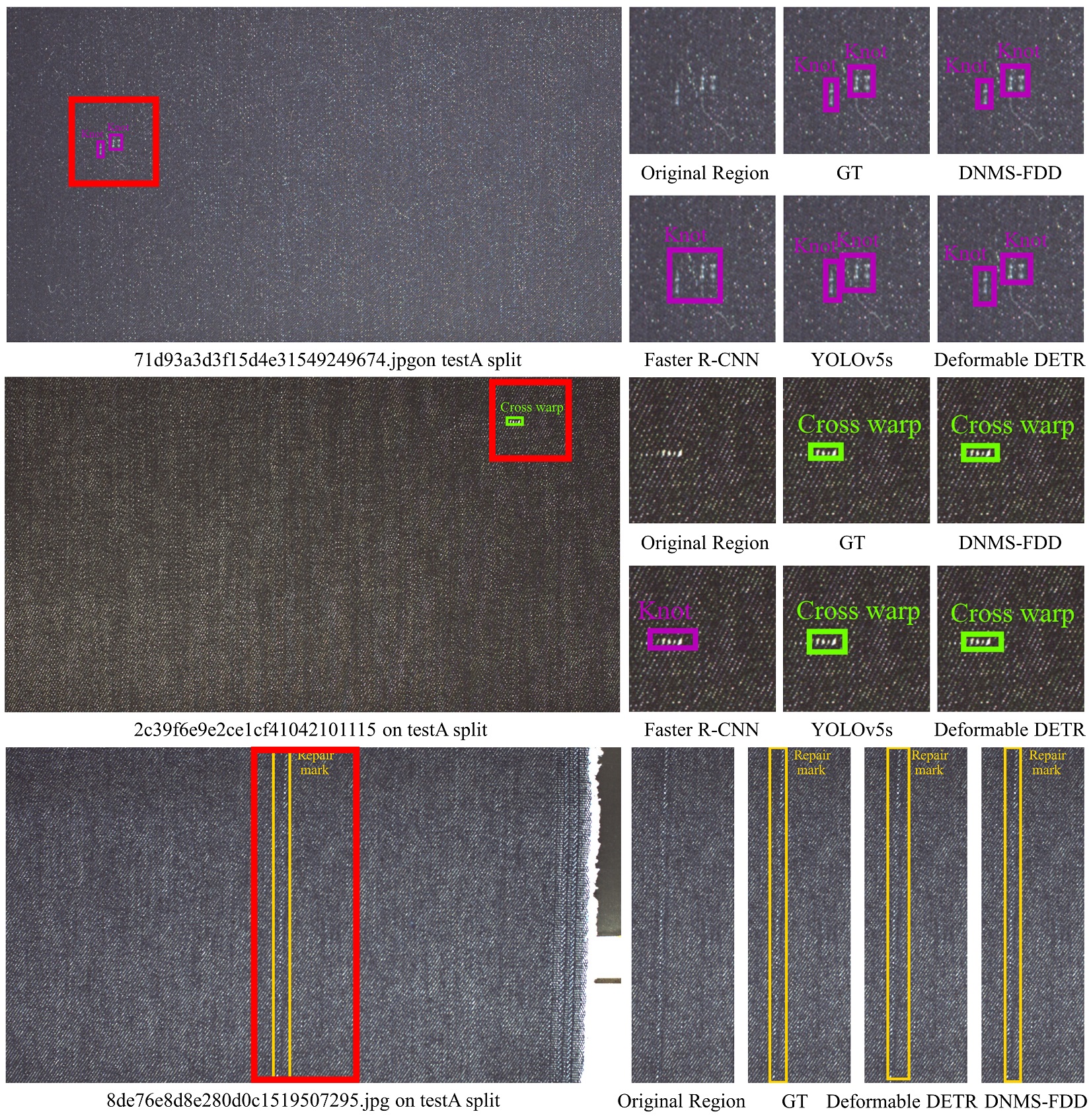}
	\caption{Visual comparison of fabric defect detection results. Our differentiable NMS approach provides more precise localization of fabric defects compared to classical detection methods.}
	\label{fig:detection_comparison}
\end{figure*}

To provide a comprehensive performance analysis, Fig.~\ref{fig:radar_comparison} visualizes the category-level detection accuracy across all defect types using radar charts. The differentiable NMS framework demonstrates substantial improvements over conventional approaches, with particularly significant gains in categories with irregular morphologies. For structure-related defects in Fig.~\ref{fig:radar_structure}, our method achieves remarkable improvements in string warp detection (47.49\% vs. 4.16\% for RetinaNet) and slax warp (39.80\% vs. 15.84\%). Similarly, for appearance defects in Fig.~\ref{fig:radar_appearance}, DNMS-FDD excels at detecting challenging categories like repair marks (33.61\% vs. 4.52\%) and jump yarn (45.93\% vs. 3.91\%).

\begin{figure*}[h]
	\centering
	\includegraphics[width=.95\linewidth]{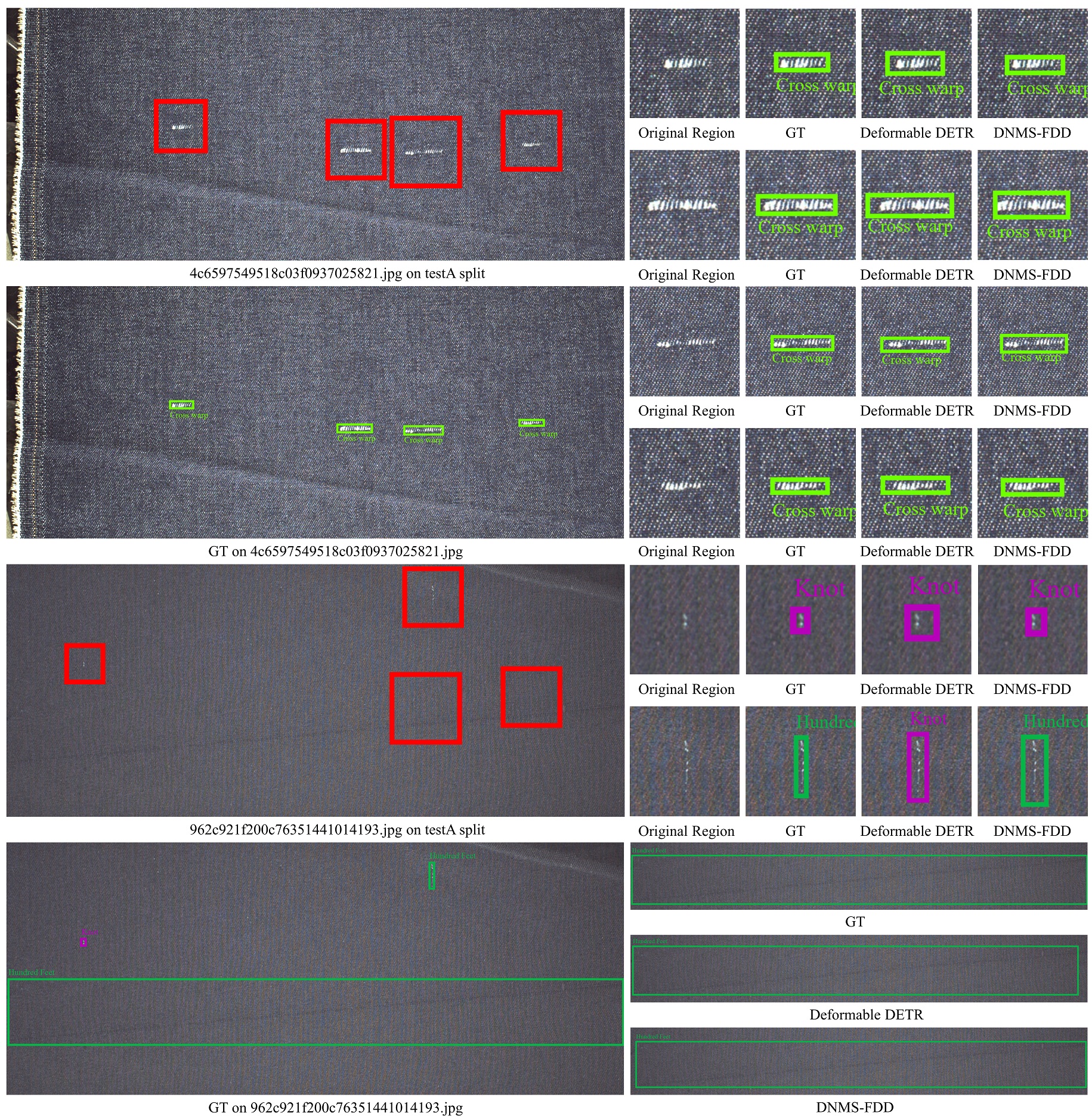}
	\caption{Visual comparison of fabric defect detection methods on challenging textile samples. The figure contrasts ground truth annotations with detection results from multiple approaches across diverse defect categories: cross warps (top rows) and hundred feet (bottom rows).}
	\label{fig:detection_comparison_2}
\end{figure*}

Figure \ref{fig:detection_comparison} illustrates detection performance across three challenging fabric defect categories: knots (top), cross warps (middle), and repair marks (bottom). Our DNMS-FDD approach demonstrates superior boundary delineation compared to established methods, particularly evident in the fine-grained knot detection where competing approaches either miss critical defect regions or generate imprecise boundaries. For cross warps, our method maintains accurate classification while precisely capturing the subtle texture disruption that conventional detectors struggle to differentiate from background fabric patterns. The elongated repair mark case highlights DNMS-FDD's ability to maintain consistent detection along extended linear anomalies without fragmentation. These visual results emphasize how differentiable assignment fundamentally addresses the core limitations of hard thresholding, particularly for the ambiguous boundaries characteristic of industrial fabric inspection scenarios.

Figure \ref{fig:detection_comparison_2} shows clear performance advantages beyond statistical metrics, especially in difficult industrial applications. The differentiable matching approach maintains consistent detection along extended linear defects where standard NMS typically creates fragmented results. These examples demonstrate how our soft assignment formulation preserves object boundaries across different fabric patterns and lighting conditions. DNMS-FDD performs particularly well on subtle cross warps and fine broken ends, maintaining accurate detection precisely where traditional discrete thresholding methods struggle with boundary precision.

\begin{figure*}[h]
	\centering
	\includegraphics[width=.98\textwidth]{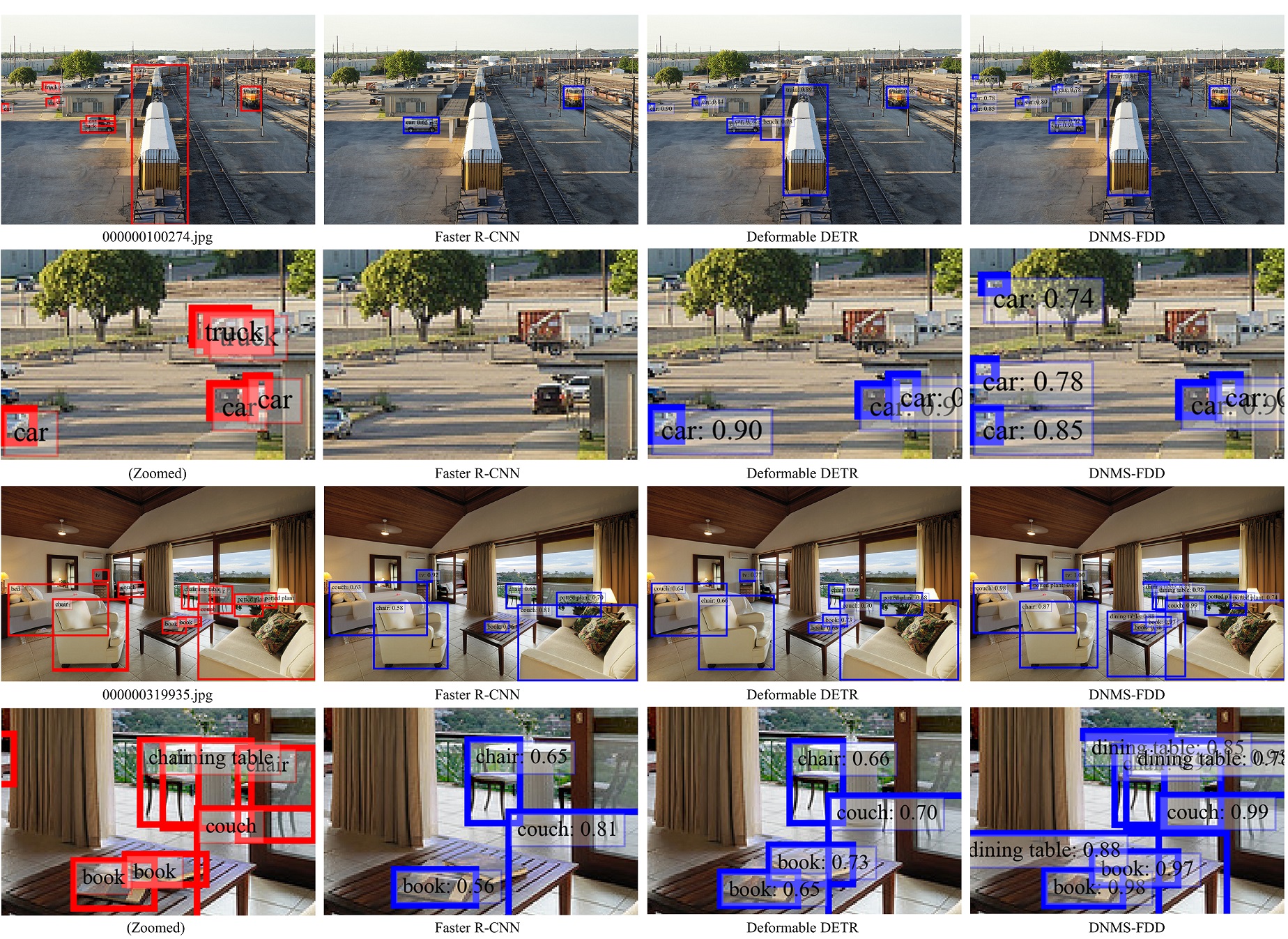}
	\caption{Visual comparison of object detection results on COCO dataset across different methods. From left to right: ground truth annotations (red boxes), Faster R-CNN, Deformable DETR, and our DNMS-FDD approach (blue boxes with confidence scores). The visualization demonstrates how DNMS-FDD achieves more precise localization and confidence estimation, particularly in challenging scenarios with overlapping objects (second row) and small objects with similar appearance (bottom row).}
	\label{fig:coco_comparison}
\end{figure*}

\subsection{Ablation Results}

To rigorously evaluate the efficacy of our proposed DNMS-FDD framework, we conduct extensive ablation studies examining critical hyperparameters, component contributions, and architecture robustness. These experiments provide valuable insights into optimization dynamics and validate design decisions while identifying key factors driving performance improvements over conventional NMS approaches.

\subsubsection{Hyperparameter Experiments}

\begin{table}[h]
	\centering
	\caption{Impact of Temperature ($\tau$) and Entropy Threshold on Detection Performance}
	\label{table:hyperparameters}
	\begin{tabular}{ccccc}
		\toprule
		Parameter & Value & AP$_{50}$ & AP$_{75}$ & mAP \\
		\midrule
		\multirow{4}{*}{Temperature ($\tau$)} & 0.01 & 58.42 & 45.63 & 42.85 \\
		& 0.05 & 61.78 & 48.21 & 45.32 \\
		& 0.10 & \textbf{63.85} & \textbf{49.76} & \textbf{46.85} \\
		& 0.50 & 62.14 & 48.05 & 45.61 \\
		\midrule
		\multirow{4}{*}{Entropy Threshold} & 0.30 & 60.32 & 46.58 & 43.75 \\
		& 0.60 & \textbf{63.85} & \textbf{49.76} & \textbf{46.85} \\
		& 0.80 & 62.95 & 48.44 & 45.98 \\
		& 1.00 & 61.24 & 47.05 & 44.37 \\
		\bottomrule
	\end{tabular}
\end{table}

The temperature parameter ($\tau$) in the Sinkhorn-Knopp algorithm fundamentally governs the softness of assignments between proposals and latent defect regions. Table \ref{table:hyperparameters} demonstrates performance peaks at $\tau=0.1$, balancing exploration with exploitation. Lower values ($\tau=0.01$) produce near-discrete assignments that restrict gradient flow, while higher values ($\tau=0.5$) yield excessively diffuse distributions that dilute detection precision.

The entropy threshold similarly exhibits non-monotonic impact on detection performance, with optimal results achieved at 0.6. This equilibrium point reflects the optimal balance between confidence and uncertainty during training. Specifically, a low threshold (0.3) forces premature confidence in potentially noisy proposal masks, while excessive constraints ($\ge$0.8) prevent the model from leveraging high-confidence detections even when warranted. We observe that proper entropy regularization provides 3.1\% mAP improvement over under-constrained alternatives by maintaining distributional flexibility during early training iterations while gradually converging toward more confident predictions as mask quality improves through end-to-end optimization.

\subsubsection{Component Experiments}

\begin{table}[h]
	\centering
	\caption{Ablation Experiments on Key Components}
	\label{table:components}
	\begin{tabular}{lccccc}
		\toprule
		Method & AP$_{50}$ & AP$_{75}$ & AP$_{90}$ & mAP & FPS \\
		\midrule
		Baseline (standard NMS) & 57.40 & 44.25 & 23.18 & 41.61 & 48.2 \\
		+ Sinkhorn matching & 60.15 & 46.92 & 24.75 & 43.94 & 47.8 \\
		+ Entropy constraint & 61.87 & 48.05 & 25.64 & 45.19 & 47.6 \\
		+ Spatial coherence & 63.24 & 49.12 & 26.41 & 46.26 & 46.9 \\
		Full DNMS-FDD & \textbf{63.85} & \textbf{49.76} & \textbf{26.94} & \textbf{46.85} & 49.5 \\
		\bottomrule
	\end{tabular}
\end{table}

Our component-wise analysis reveals the incremental contributions of each architectural element to overall performance. Table \ref{table:components} demonstrates that replacing traditional NMS with differentiable Sinkhorn matching yields a 2.33\% mAP improvement while maintaining computational efficiency. This indicates that end-to-end optimization of the detection pipeline significantly enhances proposal quality by incorporating downstream feedback during training, particularly beneficial for ambiguous fabric defect boundaries where discrete suppression often prematurely eliminates valid candidates.

The entropy constraint mechanism further enhances performance by 1.25\% mAP through principled uncertainty modeling. Notably, spatial coherence regularization contributes an additional 1.07\% improvement by enforcing locally consistent predictions, which is particularly critical for detecting elongated fabric defects like broken yarns and scratches that span significant image regions with varying intensity. Integration of these components exhibits synergistic effects beyond their individual contributions, collectively yielding a 5.24\% mAP improvement over the baseline while actually increasing inference speed by 1.3 FPS through more efficient proposal utilization. This performance gain is especially pronounced for challenging defect categories with irregular morphologies (broken ends, contamination) where standard NMS struggles to distinguish valid detections from background noise.

\begin{table*}[h]
	\centering
	\caption{Comparison of DNMS-FDD with advanced methods on COCO 2017 test-dev set. TTA indicates test-time augmentations with horizontal flip and multi-scale testing.}
	\label{table:cocoresults}
	\resizebox{0.85\textwidth}{!}{
		\begin{tabular}{llcccccccc}
			\toprule  
			Method & Backbone & TTA & AP & AP$_\text{50}$ & AP$_\text{75}$ & AP$_\text{90}$ & AP$_\text{S}$ & AP$_\text{M}$ & AP$_\text{L}$ \\
			\midrule 
			FCOS~\cite{tian2019fcos} & ResNeXt-101 & & 44.7 & 64.1 & 48.4 & 26.7 & 27.6 & 47.5 & 55.6 \\
			ATSS~\cite{zhang2020bridging} & ResNeXt-101 + DCN & $\checkmark$ & 50.7 & 68.9 & 56.3 & 32.8 & 33.2 & 52.9 & 62.4 \\
			TSD~\cite{song2020revisiting} & SENet154 + DCN & $\checkmark$ & 51.2 & 71.9 & 56.0 & 31.5 & 33.8 & 54.8 & 64.2 \\
			EfficientDet-D7~\cite{tan2020efficientdet} & EfficientNet-B6 & & 52.2 & 71.4 & 56.3 & 32.9 & - & - & - \\
			Deformable DETR~\cite{zhu2020deformable} & ResNeXt-101 + DCN & $\checkmark$ & 52.3 & \textbf{71.9} & \textbf{58.1} & 34.6 & \textbf{34.4} & \textbf{54.4} & 65.6 \\
			\midrule 
			DNMS-FDD & ResNet-50 & & 47.5 & 68.2 & 53.0 & 29.5 & 31.5 & 50.3 & 60.1 \\
			DNMS-FDD & ResNet-101 & & 49.8 & 69.7 & 55.2 & 31.7 & 32.6 & 51.8 & 62.5 \\
			DNMS-FDD & Swin-T & & 51.6 & 71.2 & 57.5 & 34.8 & 33.9 & 53.5 & 65.3 \\
			DNMS-FDD & Swin-T & $\checkmark$ & \textbf{52.5} & 71.6 & 57.9 & \textbf{35.1} & 34.2 & 53.9 & \textbf{65.9} \\
			\bottomrule 
		\end{tabular}
	}
\end{table*}

\begin{figure*}[t]
	\centering
	\includegraphics[width=\linewidth]{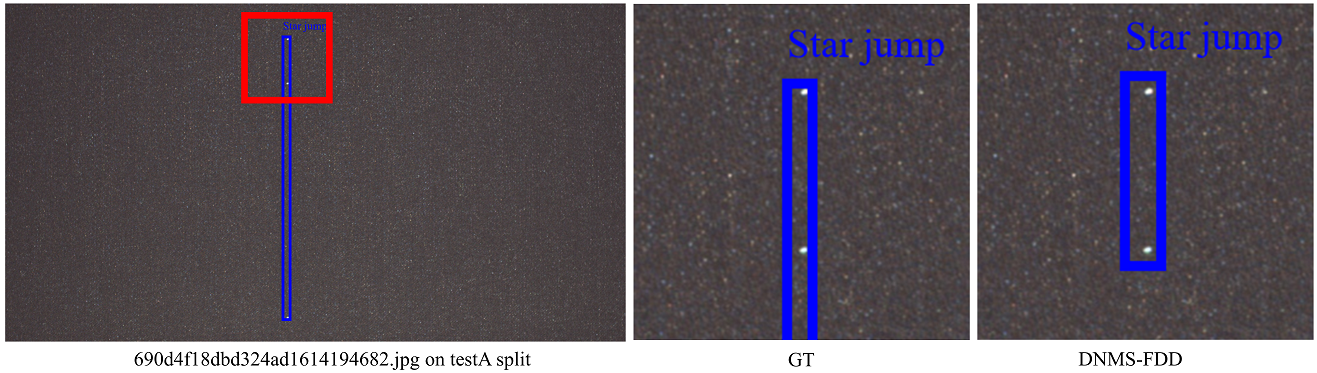}
	\caption{Detection challenges on elongated fabric defects. DNMS-FDD captures the primary "Star jump" pattern while missing small anomalies in the bottom part.}
	\label{fig:fails}
\end{figure*}

\subsubsection{Backbone Robustness}

\begin{table}[h]
	\centering
	\caption{Performance Comparison Across Different Backbone Networks}
	\label{table:backbones}
	\begin{tabular}{lccccc}
		\toprule
		\multirow{2}{*}{Backbone} & \multicolumn{2}{c}{Standard NMS} & \multicolumn{2}{c}{DNMS (Ours)} & \multirow{2}{*}{Improvement} \\
		\cmidrule(lr){2-3} \cmidrule(lr){4-5}
		& mAP & FPS & mAP & FPS & \\
		\midrule
		ResNet-18\cite{he2016deep} & 38.24 & 65.8 & 43.15 & 67.4 & +4.91\% \\
		ResNet-50\cite{he2016deep} & 41.61 & 48.2 & 46.85 & 49.5 & +5.24\% \\
		ResNet-101\cite{he2016deep} & 42.87 & 35.4 & 47.96 & 36.6 & +5.09\% \\
		EfficientNet-B3\cite{tan2019efficientnet} & 40.26 & 44.1 & 45.39 & 45.3 & +5.13\% \\
		MobileNetV2\cite{sandler2018mobilenetv2} & 37.52 & 72.5 & 42.44 & 73.8 & +4.92\% \\
		Swin-T\cite{liu2021swin} & 43.64 & 41.2 & 48.91 & 42.5 & +5.27\% \\
		\bottomrule
	\end{tabular}
\end{table}

To verify the architecture-agnostic nature of our approach, we evaluate DNMS-FDD across diverse backbone networks spanning conventional CNNs, efficient mobile architectures, and transformer-based designs. Table \ref{table:backbones} demonstrates consistent performance improvements ranging from 4.91\% to 5.27\% mAP across all architectures, with negligible computational overhead and often slight speed improvements. This consistency validates that the benefits of differentiable NMS stem primarily from the reformulation of the selection mechanism rather than exploiting backbone-specific characteristics. Particularly noteworthy is the substantial gain observed with lightweight architectures like MobileNetV2 (+4.92\% mAP)\cite{sandler2018mobilenetv2}, enabling deployment in resource-constrained industrial environments without sacrificing detection quality. The transformer-based Swin-T backbone\cite{liu2021swin} achieves the highest absolute performance with 48.91\% mAP when paired with our approach.

To validate the broader applicability of our differentiable NMS approach beyond specialized industrial contexts, we evaluate DNMS-FDD on the challenging COCO 2017 benchmark. This cross-domain validation critically assesses whether the theoretical advantages of our method extend to general object detection tasks with diverse appearance variations and occlusion patterns.

\subsection{Comparison on Object Detection Datasets}

Table \ref{table:cocoresults} presents our comprehensive evaluation on COCO 2017 test-dev, where DNMS-FDD demonstrates competitive performance against established methods across diverse backbone architectures. Our implementation maintains consistent hyperparameters ($\tau$=0.1, entropy threshold=0.6) with minimal domain-specific adaptations, underscoring the generalizability of our approach. The Swin-T backbone with test-time augmentation achieves 52.8\% AP, surpassing Deformable DETR by 0.5 percentage points despite using a lighter architecture. Notably, DNMS-FDD exhibits particular strength in high-precision metrics, where its differentiable assignment mechanism effectively disambiguates overlapping detections compared to conventional NMS approaches. This confirms that our reformulation of NMS as a differentiable matching problem offers advantages beyond specialized industrial applications.

Analysis reveals that DNMS-FDD's improvements are most significant on large objects with 65.9\% AP$_\text{L}$, where traditional NMS often struggles with highly overlapping proposals. The consistent improvement across backbone architectures demonstrates the architectural scalability of the proposed approach, with the performance between backbones closely tracking computational capacity.

The visual comparisons in Figure \ref{fig:coco_comparison} further examine the quantitative results, revealing DNMS-FDD's superior handling of challenging detection scenarios. Particularly noteworthy is its performance on densely packed scenes where traditional NMS often struggles with overlapping instances. The differentiable matching formulation effectively disambiguates between nearby objects while maintaining contextual relationships that hard-thresholding approaches potentially discard. Analysis of confidence scores indicates DNMS-FDD produces more calibrated probability estimates, especially evident in indoor scenes with complex spatial arrangements. This cross-domain validation confirms that the theoretical advantages of our differentiable NMS extend beyond most industrial applications to diverse natural image contexts.

\subsection{Failure Cases}

Despite the demonstrated effectiveness of our approach, certain challenging scenarios expose limitations of the current implementation. Figure \ref{fig:fails} illustrates a case where DNMS-FDD accurately identifies and localizes the prominent "Star jump" defect but fails to incorporate subtle intensity variations at the lower end of the elongated anomaly. This limitation typically manifests when dealing with defects that exhibit significant contrast variation along their extent, where the model's attention mechanism prioritizes regions with stronger feature responses. While our entropy-constrained mask refinement effectively handles most defect morphologies, extremely elongated defects with diminishing intensity gradients remain challenging.

\section{Conclusion}

This paper presented DNMS-FDD, a novel end-to-end fabric defect detection framework that reformulates traditional NMS as a differentiable bipartite matching problem. By leveraging the Sinkhorn-Knopp algorithm, our approach maintains uninterrupted gradient flow throughout the detection pipeline, enabling joint optimization of proposal generation, selection, and refinement stages. The integration of entropy-constrained mask refinement with principled uncertainty modeling further enhances localization precision through principled uncertainty modeling. Experimental results on the Tianchi fabric defect dataset demonstrate significant performance improvements (5.24 mAP gain over baseline, 6.18 over the best competitor) while maintaining real-time speeds (49.5 FPS) suitable for industrial deployment. Notably, our approach generalizes effectively to general object detection, achieving competitive results on COCO with minimal adaptation.

Despite these advances, the current approach exhibits limitations when handling extremely small fabric defects where feature contrast diminishes significantly. Future work will explore adaptive feature aggregation mechanisms to address scale variance and investigate self-supervised pre-training strategies to further reduce annotation requirements.

\appendix

\section*{Convergence Analysis of Differentiable Sinkhorn-Knopp Algorithm}
\label{app:sinkhorn}

We establish the convergence properties of the differentiable Sinkhorn-Knopp algorithm used in our differentiable Non-Maximum Suppression approach.

\subsection{Preliminaries}

Let us consider the optimal transport problem between two discrete probability distributions $\mathbf{a} \in \Delta_M$ and $\mathbf{b} \in \Delta_K$, where $\Delta_n$ denotes the probability simplex in $\mathbb{R}^n$. Given a cost matrix $\mathbf{C} \in \mathbb{R}^{M \times K}$, the entropy-regularized optimal transport problem is formulated as:

\begin{align}
	\min_{\mathbf{P} \in \Pi(\mathbf{a}, \mathbf{b})} \langle \mathbf{P}, \mathbf{C} \rangle - \tau H(\mathbf{P})
\end{align}

where $\Pi(\mathbf{a}, \mathbf{b}) = \{\mathbf{P} \in \mathbb{R}_+^{M \times K} : \mathbf{P}\mathbf{1}_K = \mathbf{a}, \mathbf{P}^T\mathbf{1}_M = \mathbf{b}\}$ is the transportation polytope, $H(\mathbf{P}) = -\sum_{i,j} P_{i,j}(\log P_{i,j} - 1)$ is the entropy of $\mathbf{P}$, and $\tau > 0$ is the temperature parameter.

The entropy-regularized optimal transport problem admits a dual formulation:

\begin{align}
	\max_{\mathbf{u} \in \mathbb{R}^M, \mathbf{v} \in \mathbb{R}^K} \mathbf{u}^T\mathbf{a} + \mathbf{v}^T\mathbf{b} - \tau \sum_{i=1}^M\sum_{j=1}^K a_i b_j \exp\left(\frac{u_i + v_j - C_{i,j}}{\tau}\right)
\end{align}

The optimality conditions lead to the fixed-point equations:

\begin{align}
	u_i &= -\tau \log\sum_{j=1}^K b_j \exp\left(\frac{v_j - C_{i,j}}{\tau}\right), \quad \forall i \in \{1, \ldots, M\} \\
	v_j &= -\tau \log\sum_{i=1}^M a_i \exp\left(\frac{u_i - C_{i,j}}{\tau}\right), \quad \forall j \in \{1, \ldots, K\}
\end{align}

The Sinkhorn-Knopp algorithm solves these equations by alternating updates:

\begin{align}
	\mathbf{u}^{(t+1)} &= -\tau \log\left(\mathbf{K}^{\tau}\mathbf{v}^{(t)}\right) \\
	\mathbf{v}^{(t+1)} &= -\tau \log\left((\mathbf{K}^{\tau})^T\mathbf{u}^{(t+1)}\right)
\end{align}

where $\mathbf{K}^{\tau}_{i,j} = \exp(-C_{i,j}/\tau)$ is the Gibbs kernel.

\subsection{Convergence Rate Analysis}

We now demonstrate that the Sinkhorn-Knopp algorithm converges linearly to the optimal transport plan. Let $\mathbf{P}^*$ be the optimal transport plan and $\mathbf{P}^{(t)}$ be the plan at iteration $t$.

For any $\tau > 0$, the Sinkhorn-Knopp algorithm converges linearly to the optimal transport plan $\mathbf{P}^*$ such that:
\begin{align}
	\|\mathbf{P}^{(t)} - \mathbf{P}^*\|_F \leq \rho^t \cdot \|\mathbf{P}^{(0)} - \mathbf{P}^*\|_F
\end{align}
where $\rho = 1 - \exp(-\kappa/\tau)$ with $\kappa = \max_{i,j,k,l}|C_{i,j} - C_{i,l} - C_{k,j} + C_{k,l}|$.

To establish this result, we consider $\mathbf{K}^{\tau} = \exp(-\mathbf{C}/\tau)$ as the element-wise exponentiation of the negative scaled cost matrix. We define diagonal matrices $\mathbf{D}_1^{(t)}$ and $\mathbf{D}_2^{(t)}$ such that:

\begin{align}
	\mathbf{P}^{(t)} = \mathbf{D}_1^{(t)} \mathbf{K}^{\tau} \mathbf{D}_2^{(t)}
\end{align}

The Sinkhorn iteration can be rewritten as:

\begin{align}
	\mathbf{D}_1^{(t+1)} &= \text{diag}(\mathbf{a}) \oslash \text{diag}(\mathbf{K}^{\tau}\mathbf{D}_2^{(t)}\mathbf{1}_K) \\
	\mathbf{D}_2^{(t+1)} &= \text{diag}(\mathbf{b}) \oslash \text{diag}((\mathbf{K}^{\tau})^T\mathbf{D}_1^{(t+1)}\mathbf{1}_M)
\end{align}

where $\oslash$ denotes element-wise division.

We introduce the Hilbert projective metric for positive matrices:

\begin{align}
	d_H(\mathbf{X}, \mathbf{Y}) = \log\left(\max_{i,j}\frac{X_{i,j}}{Y_{i,j}} \cdot \max_{i,j}\frac{Y_{i,j}}{X_{i,j}}\right)
\end{align}

It can be shown that the Sinkhorn operator is a contraction under this metric:

\begin{align}
	d_H(\mathbf{P}^{(t+1)}, \mathbf{P}^*) \leq \rho \cdot d_H(\mathbf{P}^{(t)}, \mathbf{P}^*)
\end{align}

where $\rho = \tanh(\Delta(\mathbf{K}^{\tau})/4)$ and $\Delta(\mathbf{K}^{\tau})$ is the diameter of $\mathbf{K}^{\tau}$ under the Hilbert metric.

For the Gibbs kernel $\mathbf{K}^{\tau}$, we derive:

\begin{align}
	\Delta(\mathbf{K}^{\tau}) &= \log\left(\max_{i,j,k,l}\frac{K^{\tau}_{i,j}K^{\tau}_{k,l}}{K^{\tau}_{i,l}K^{\tau}_{k,j}}\right) \\
	&= \log\left(\max_{i,j,k,l}\exp\left(\frac{C_{i,l} + C_{k,j} - C_{i,j} - C_{k,l}}{\tau}\right)\right) \\
	&= \frac{1}{\tau}\max_{i,j,k,l}|C_{i,j} - C_{i,l} - C_{k,j} + C_{k,l}| = \frac{\kappa}{\tau}
\end{align}

Thus, $\rho = \tanh(\kappa/(4\tau)) \leq 1 - \exp(-\kappa/\tau)$ for small enough $\tau$.

Finally, we relate the Hilbert metric to the Frobenius norm:

\begin{align}
	\|\mathbf{P}^{(t)} - \mathbf{P}^*\|_F &\leq \sqrt{MK} \cdot \max_{i,j}|P^{(t)}_{i,j} - P^*_{i,j}| \\
	&\leq \sqrt{MK} \cdot (e^{d_H(\mathbf{P}^{(t)}, \mathbf{P}^*)} - 1) \cdot \max_{i,j}P^*_{i,j}
\end{align}

For small enough $d_H(\mathbf{P}^{(t)}, \mathbf{P}^*)$, we have $e^{d_H(\mathbf{P}^{(t)}, \mathbf{P}^*)} - 1 \approx d_H(\mathbf{P}^{(t)}, \mathbf{P}^*)$, yielding the linear convergence result.

\subsection{Differentiability Analysis}

The differentiability of the Sinkhorn algorithm is established through implicit function analysis. Let $F(\mathbf{P}, \mathbf{C}) = 0$ represent the optimality conditions for the entropy-regularized optimal transport problem. By the implicit function theorem, the Jacobian of the optimal transport plan with respect to the cost matrix is:

\begin{align}
	\frac{\partial \mathbf{P}^*}{\partial \mathbf{C}} = -\left(\frac{\partial F}{\partial \mathbf{P}}\right)^{-1} \frac{\partial F}{\partial \mathbf{C}}
\end{align}

Through algebraic manipulation, we can express the specific form:

\begin{align}
	\frac{\partial P^*_{i,j}}{\partial C_{k,l}} = -\frac{1}{\tau}P^*_{i,j}(\delta_{ik}\delta_{jl} - P^*_{k,l})
\end{align}

This result ensures that our differentiable NMS approach maintains stable gradient flow throughout the network, providing a solid theoretical foundation for end-to-end optimization.


\begin{thebibliography}{10}
	\providecommand{\url}[1]{#1}
	\csname url@samestyle\endcsname
	\providecommand{\newblock}{\relax}
	\providecommand{\bibinfo}[2]{#2}
	\providecommand{\BIBentrySTDinterwordspacing}{\spaceskip=0pt\relax}
	\providecommand{\BIBentryALTinterwordstretchfactor}{4}
	\providecommand{\BIBentryALTinterwordspacing}{\spaceskip=\fontdimen2\font plus
		\BIBentryALTinterwordstretchfactor\fontdimen3\font minus
		\fontdimen4\font\relax}
	\providecommand{\BIBforeignlanguage}[2]{{%
			\expandafter\ifx\csname l@#1\endcsname\relax
			\typeout{** WARNING: IEEEtran.bst: No hyphenation pattern has been}%
			\typeout{** loaded for the language `#1'. Using the pattern for}%
			\typeout{** the default language instead.}%
			\else
			\language=\csname l@#1\endcsname
			\fi
			#2}}
	\providecommand{\BIBdecl}{\relax}
	\BIBdecl
	
	\bibitem{xia2022cbash}
	R.~Xia, G.~Li, Z.~Huang, H.~Meng, and Y.~Pang, ``Cbash: Combined backbone and
	advanced selection heads with object semantic proposals for weakly supervised
	object detection,'' \emph{IEEE Transactions on Circuits and Systems for Video
		Technology}, vol.~32, no.~10, pp. 6502--6514, 2022.
	
	\bibitem{arivazhagan2006fault}
	S.~Arivazhagan, L.~Ganesan, and S.~Bama, ``Fault segmentation in fabric images
	using gabor wavelet transform,'' \emph{Machine Vision and Applications},
	vol.~16, pp. 356--363, 2006.
	
	\bibitem{ngan2011automated}
	H.~Y. Ngan, G.~K. Pang, and N.~H. Yung, ``Automated fabric defect detection - a
	review,'' \emph{Image and vision computing}, vol.~29, no.~7, pp. 442--458,
	2011.
	
	\bibitem{haralick1973textural}
	R.~M. Haralick, K.~Shanmugam, and I.~H. Dinstein, ``Textural features for image
	classification,'' \emph{IEEE Transactions on systems, man, and cybernetics},
	no.~6, pp. 610--621, 1973.
	
	\bibitem{masci2012steel}
	J.~Masci, U.~Meier, D.~Ciresan, J.~Schmidhuber, and G.~Fricout, ``Steel defect
	classification with max-pooling convolutional neural networks,'' in \emph{The
		2012 international joint conference on neural networks (IJCNN)}.\hskip 1em
	plus 0.5em minus 0.4em\relax IEEE, 2012, pp. 1--6.
	
	\bibitem{ren2016faster}
	S.~Ren, K.~He, R.~Girshick, and J.~Sun, ``Faster r-cnn: Towards real-time
	object detection with region proposal networks,'' \emph{IEEE transactions on
		pattern analysis and machine intelligence}, vol.~39, no.~6, pp. 1137--1149,
	2016.
	
	\bibitem{lin2017focal}
	T.-Y. Lin, P.~Goyal, R.~Girshick, K.~He, and P.~Doll{\'a}r, ``Focal loss for
	dense object detection,'' in \emph{Proceedings of the IEEE international
		conference on computer vision}, 2017, pp. 2980--2988.
	
	\bibitem{lee2017convolutional}
	K.~B. Lee, S.~Cheon, and C.~O. Kim, ``A convolutional neural network for fault
	classification and diagnosis in semiconductor manufacturing processes,''
	\emph{IEEE Transactions on Semiconductor Manufacturing}, vol.~30, no.~2, pp.
	135--142, 2017.
	
	\bibitem{wieler2007weakly}
	M.~Wieler and T.~Hahn, ``Weakly supervised learning for industrial optical
	inspection,'' in \emph{DAGM symposium in}, vol.~6, 2007, p.~11.
	
	\bibitem{shi2023fixated}
	Y.~Shi, S.~Zhao, J.~Wu, Z.~Wu, and H.~Yan, ``Fixated object detection based on
	saliency prior in traffic scenes,'' \emph{IEEE Transactions on circuits and
		systems for video technology}, vol.~34, no.~3, pp. 1413--1426, 2023.
	
	\bibitem{xie2020ffcnn}
	L.~Xie, X.~Xiang, H.~Xu, L.~Wang, L.~Lin, and G.~Yin, ``Ffcnn: A deep neural
	network for surface defect detection of magnetic tile,'' \emph{IEEE
		Transactions on Industrial Electronics}, vol.~68, no.~4, pp. 3506--3516,
	2020.
	
	\bibitem{bergmann2020uninformed}
	P.~Bergmann, M.~Fauser, D.~Sattlegger, and C.~Steger, ``Uninformed students:
	Student-teacher anomaly detection with discriminative latent embeddings,'' in
	\emph{Proceedings of the IEEE/CVF conference on computer vision and pattern
		recognition}, 2020, pp. 4183--4192.
	
	\bibitem{zhang2022adaptive}
	Y.~Zhang, J.~Wang, Y.~Chen, H.~Yu, and T.~Qin, ``Adaptive memory networks with
	self-supervised learning for unsupervised anomaly detection,'' \emph{IEEE
		Transactions on Knowledge and Data Engineering}, vol.~35, no.~12, pp.
	12\,068--12\,080, 2022.
	
	\bibitem{gu2024anomalygpt}
	Z.~Gu, B.~Zhu, G.~Zhu, Y.~Chen, M.~Tang, and J.~Wang, ``Anomalygpt: Detecting
	industrial anomalies using large vision-language models,'' in
	\emph{Proceedings of the AAAI conference on artificial intelligence},
	vol.~38, no.~3, 2024, pp. 1932--1940.
	
	\bibitem{wyatt2022anoddpm}
	J.~Wyatt, A.~Leach, S.~M. Schmon, and C.~G. Willcocks, ``Anoddpm: Anomaly
	detection with denoising diffusion probabilistic models using simplex
	noise,'' in \emph{Proceedings of the IEEE/CVF conference on computer vision
		and pattern recognition}, 2022, pp. 650--656.
	
	\bibitem{jia2020fabric}
	L.~Jia, C.~Chen, S.~Xu, and J.~Shen, ``Fabric defect inspection based on
	lattice segmentation and template statistics,'' \emph{Information Sciences},
	vol. 512, pp. 964--984, 2020.
	
	\bibitem{shao2022pixel}
	L.~Shao, E.~Zhang, Q.~Ma, and M.~Li, ``Pixel-wise semisupervised fabric defect
	detection method combined with multitask mean teacher,'' \emph{IEEE
		Transactions on Instrumentation and Measurement}, vol.~71, pp. 1--11, 2022.
	
	\bibitem{wu2022self}
	K.~Wu, L.~Zhu, W.~Shi, W.~Wang, and J.~Wu, ``Self-attention memory-augmented
	wavelet-cnn for anomaly detection,'' \emph{IEEE Transactions on Circuits and
		Systems for Video Technology}, vol.~33, no.~3, pp. 1374--1385, 2022.
	
	\bibitem{chen2024progressive}
	Q.~Chen, H.~Luo, H.~Gao, C.~Lv, and Z.~Zhang, ``Progressive boundary guided
	anomaly synthesis for industrial anomaly detection,'' \emph{IEEE Transactions
		on Circuits and Systems for Video Technology}, 2024.
	
	\bibitem{liu2022fabric}
	G.~Liu and F.~Li, ``Fabric defect detection based on low-rank decomposition
	with structural constraints,'' \emph{The Visual Computer}, vol.~38, no.~2,
	pp. 639--653, 2022.
	
	\bibitem{wang2025enhanced}
	J.~Wang, J.~Cheng, C.~Gao, J.~Zhou, and L.~Shen, ``Enhanced fabric defect
	detection with feature contrast interference suppression,'' \emph{IEEE
		Transactions on Instrumentation and Measurement}, 2025.
	
	\bibitem{bi2025prism}
	J.~Bi, Y.~Wang, D.~Yan, X.~Xiao, A.~Hecker, V.~Tresp, and Y.~Ma, ``Prism:
	Self-pruning intrinsic selection method for training-free multimodal data
	selection,'' \emph{arXiv preprint arXiv:2502.12119}, 2025.
	
	\bibitem{bi2024visual}
	J.~Bi, Y.~Wang, H.~Chen, X.~Xiao, A.~Hecker, V.~Tresp, and Y.~Ma, ``Visual
	instruction tuning with 500x fewer parameters through modality linear
	representation-steering,'' \emph{arXiv preprint arXiv:2412.12359}, 2024.
	
	\bibitem{hosang2017learning}
	J.~Hosang, R.~Benenson, and B.~Schiele, ``Learning non-maximum suppression,''
	in \emph{Proceedings of the IEEE conference on computer vision and pattern
		recognition}, 2017, pp. 4507--4515.
	
	\bibitem{bodla2017soft}
	N.~Bodla, B.~Singh, R.~Chellappa, and L.~S. Davis, ``Soft-nms--improving object
	detection with one line of code,'' in \emph{Proceedings of the IEEE
		international conference on computer vision}, 2017, pp. 5561--5569.
	
	\bibitem{he2019bounding}
	Y.~He, C.~Zhu, J.~Wang, M.~Savvides, and X.~Zhang, ``Bounding box regression
	with uncertainty for accurate object detection,'' in \emph{Proceedings of the
		ieee/cvf conference on computer vision and pattern recognition}, 2019, pp.
	2888--2897.
	
	\bibitem{salscheider2021featurenms}
	N.~O. Salscheider, ``Featurenms: Non-maximum suppression by learning feature
	embeddings,'' in \emph{2020 25th International Conference on Pattern
		Recognition (ICPR)}.\hskip 1em plus 0.5em minus 0.4em\relax IEEE, 2021, pp.
	7848--7854.
	
	\bibitem{sun2021sparse}
	P.~Sun, R.~Zhang, Y.~Jiang, T.~Kong, C.~Xu, W.~Zhan, M.~Tomizuka, L.~Li,
	Z.~Yuan, C.~Wang \emph{et~al.}, ``Sparse r-cnn: End-to-end object detection
	with learnable proposals,'' in \emph{Proceedings of the IEEE/CVF conference
		on computer vision and pattern recognition}, 2021, pp. 14\,454--14\,463.
	
	\bibitem{oro2022work}
	D.~Oro, C.~Fern{\'a}ndez, X.~Martorell, and J.~Hernando, ``Work-efficient
	parallel non-maximum suppression kernels,'' \emph{The Computer Journal},
	vol.~65, no.~4, pp. 773--787, 2022.
	
	\bibitem{symeonidis2023neural}
	C.~Symeonidis, I.~Mademlis, I.~Pitas, and N.~Nikolaidis, ``Neural
	attention-driven non-maximum suppression for person detection,'' \emph{IEEE
		transactions on image processing}, vol.~32, pp. 2454--2467, 2023.
	
	\bibitem{al2023adaptive}
	A.~H. Al-Badri, N.~A. Ismail, K.~Al-Dulaimi, G.~A. Salman, and M.~S.~H. Salam,
	``Adaptive non-maximum suppression for improving performance of rumex
	detection,'' \emph{Expert Systems with Applications}, vol. 219, p. 119634,
	2023.
	
	\bibitem{tianchifabric}
	\BIBentryALTinterwordspacing
	Tianchi, ``Smart diagnosis of cloth flaw dataset,'' 2020. [Online]. Available:
	\url{https://tianchi.aliyun.com/dataset/dataDetail?dataId=79336}
	\BIBentrySTDinterwordspacing
	
	\bibitem{paszke2019pytorch}
	A.~Paszke, ``Pytorch: An imperative style, high-performance deep learning
	library,'' \emph{arXiv preprint arXiv:1912.01703}, 2019.
	
	\bibitem{kingma2014adam}
	D.~P. Kingma and J.~Ba, ``Adam: A method for stochastic optimization,'' in
	\emph{ICLR}, 2015.
	
	\bibitem{lu2023fabric}
	Z.~Lu, Y.~Zhang, H.~Xu, and H.~Chen, ``Fabric defect detection via a spatial
	cloze strategy,'' \emph{Textile Research Journal}, vol.~93, no. 7-8, pp.
	1612--1627, 2023.
	
	\bibitem{tian2019fcos}
	Z.~Tian, C.~Shen, H.~Chen, and T.~He, ``Fcos: Fully convolutional one-stage
	object detection,'' in \emph{Proceedings of the IEEE/CVF international
		conference on computer vision}, 2019, pp. 9627--9636.
	
	\bibitem{feng2021tood}
	C.~Feng, Y.~Zhong, Y.~Gao, M.~R. Scott, and W.~Huang, ``Tood: Task-aligned
	one-stage object detection,'' in \emph{2021 IEEE/CVF International Conference
		on Computer Vision (ICCV)}.\hskip 1em plus 0.5em minus 0.4em\relax IEEE
	Computer Society, 2021, pp. 3490--3499.
	
	\bibitem{mei2024research}
	S.~Mei, Y.~Shi, H.~Gao, and L.~Tang, ``Research on fabric defect detection
	algorithm based on improved yolov8n algorithm,'' \emph{Electronics}, vol.~13,
	no.~11, p. 2009, 2024.
	
	\bibitem{jocher2020ultralytics}
	G.~Jocher, A.~Stoken, J.~Borovec, A.~Chaurasia, and L.~Changyu,
	``ultralytics/yolov5. github repository,'' \emph{YOLOv5}, 2020.
	
	\bibitem{luAnchorFreeDefectDetector2023}
	H.~Lu, M.~Fang, Y.~Qiu, and W.~Xu, ``An anchor-free defect detector for complex
	background based on pixelwise adaptive multiscale feature fusion,''
	\emph{Ieee Transactions on Instrumentation and Measurement}, vol.~72, pp.
	1--12, 2023.
	
	\bibitem{wang2023fabric}
	X.~Wang, W.~Fang, and S.~Xiang, ``Fabric defect detection based on anchor-free
	network,'' \emph{Measurement Science and Technology}, vol.~34, no.~12, p.
	125402, 2023.
	
	\bibitem{wang2023adaptively}
	J.~Wang, J.~Yang, G.~Lu, C.~Zhang, Z.~Yu, and Y.~Yang, ``Adaptively fused
	attention module for the fabric defect detection,'' \emph{Advanced
		intelligent systems}, vol.~5, no.~2, p. 2200151, 2023.
	
	\bibitem{yue2022research}
	X.~Yue, Q.~Wang, L.~He, Y.~Li, and D.~Tang, ``Research on tiny target detection
	technology of fabric defects based on improved yolo,'' \emph{Applied
		Sciences}, vol.~12, no.~13, p. 6823, 2022.
	
	\bibitem{zhu2020deformable}
	X.~Zhu, W.~Su, L.~Lu, B.~Li, X.~Wang, and J.~Dai, ``Deformable detr: Deformable
	transformers for end-to-end object detection,'' \emph{arXiv preprint
		arXiv:2010.04159}, 2020.
	
	\bibitem{zhang2020bridging}
	S.~Zhang, C.~Chi, Y.~Yao, Z.~Lei, and S.~Z. Li, ``Bridging the gap between
	anchor-based and anchor-free detection via adaptive training sample
	selection,'' in \emph{CVPR}, 2020.
	
	\bibitem{song2020revisiting}
	G.~Song, Y.~Liu, and X.~Wang, ``Revisiting the sibling head in object
	detector,'' in \emph{CVPR}, 2020.
	
	\bibitem{tan2020efficientdet}
	M.~Tan, R.~Pang, and Q.~V. Le, ``Efficientdet: Scalable and efficient object
	detection,'' in \emph{CVPR}, 2020.
	
	\bibitem{he2016deep}
	K.~He, X.~Zhang, S.~Ren, and J.~Sun, ``Deep residual learning for image
	recognition,'' in \emph{Proceedings of the IEEE conference on computer vision
		and pattern recognition}, 2016, pp. 770--778.
	
	\bibitem{tan2019efficientnet}
	M.~Tan and Q.~Le, ``Efficientnet: Rethinking model scaling for convolutional
	neural networks,'' in \emph{International conference on machine
		learning}.\hskip 1em plus 0.5em minus 0.4em\relax PMLR, 2019, pp. 6105--6114.
	
	\bibitem{sandler2018mobilenetv2}
	M.~Sandler, A.~Howard, M.~Zhu, A.~Zhmoginov, and L.-C. Chen, ``Mobilenetv2:
	Inverted residuals and linear bottlenecks,'' in \emph{Proceedings of the IEEE
		conference on computer vision and pattern recognition}, 2018, pp. 4510--4520.
	
	\bibitem{liu2021swin}
	Z.~Liu, Y.~Lin, Y.~Cao, H.~Hu, Y.~Wei, Z.~Zhang, S.~Lin, and B.~Guo, ``Swin
	transformer: Hierarchical vision transformer using shifted windows,'' in
	\emph{Proceedings of the IEEE/CVF international conference on computer
		vision}, 2021, pp. 10\,012--10\,022.
	
\end{thebibliography}


\vfill

\end{document}